\documentclass[11pt]{article}

\usepackage[preprint]{acl}

\usepackage{times}
\usepackage{latexsym}
\usepackage{amsfonts}

\usepackage[T1]{fontenc}
\usepackage[utf8]{inputenc}

\usepackage{microtype}

\usepackage[most]{tcolorbox}
\usepackage{tabularx}
\usepackage{array}
\usepackage{xcolor}
\usepackage{amsmath}

\definecolor{examplegold}{HTML}{FFF3BF}
\definecolor{exampleborder}{HTML}{7A8594}
\definecolor{exampletitle}{HTML}{EEF2F6}

\newcolumntype{Y}{>{\raggedright\arraybackslash}X}

\newtcolorbox{benchmarkcard}[1]{
enhanced,
width=\linewidth,
colback=white,
colframe=exampleborder,
colbacktitle=exampletitle,
coltitle=black,
title={#1},
fonttitle=\bfseries\small,
boxrule=0.6pt,
arc=1.5mm,
left=1.5mm,
right=1.5mm,
top=1.2mm,
bottom=1.2mm,
before skip=0pt,
after skip=0pt
}

\usepackage{etoolbox}

\newcommand{\benchmarkoption}[4]{%
\noindent
\textbf{#1.}\ #3\ \textsc{#2}\ #4
\par\smallskip
}
\newcommand{\goldbenchmarkoption}[4]{%
\noindent
\colorbox{examplegold}{%
\parbox{\dimexpr\linewidth-2\fboxsep\relax}{%
\textbf{#1.}\ #3\ \textsc{#2}\ #4
\hfill\textbf{Gold}%
}%
}%
\par\smallskip
}
\newcommand{\benchmarkoptionNN}[3]{%
\noindent
\textbf{#1.}\ Neither\ #2\ Nor\ #3
\par\smallskip
}
\newcommand{\goldbenchmarkoptionNN}[3]{%
\noindent
\colorbox{examplegold}{%
\parbox{\dimexpr\linewidth-2\fboxsep\relax}{%
\textbf{#1.}\ Neither\ #2\ Nor\ #3
\hfill\textbf{Gold}%
}%
}%
\par\smallskip
}
\usepackage{inconsolata}
\usepackage{booktabs}
\usepackage{graphicx}
\usepackage{adjustbox} 
\usepackage{amsmath}
\usepackage{multirow} 
\usepackage{enumitem}
\usepackage{tabularx}
\usepackage{float}
\usepackage{makecell}
\usepackage{amsthm}
\usepackage{amsmath}

\newcommand{\LCQA}{\textsc{Logical\mbox{-}CommonsenseQA}}
\newcommand{\LSATA}{\textsc{Logical\mbox{-}SATA}}

\title{From Atomic Evidence to Logical Composition: \\Structured Compositional Reasoning over Compound Answer Options}

\author{Obed Junias \\
  University of Colorado Boulder\\
  \texttt{obed.junias@colorado.edu} \\\And
  Maria Leonor Pacheco \\
  University of Colorado Boulder\\
  \texttt{maria.pacheco@colorado.edu} \\}

\begin{document}
\maketitle
\begin{abstract}
Large language models often fail when answer options require combining atomic judgments under explicit logical operators, even when they judge the individual atoms correctly. We study compound options connected by \textsc{And}, \textsc{Or}, and \textsc{Neither/Nor}, introducing a framework that decomposes each option into atomic answers and scores contrastive hypotheses about each one, so the model never sees a compound option. An operator-constrained integer linear program then composes the calibrated scores into a single prediction. We evaluate on \LCQA~and introduce \LSATA, a reading-comprehension benchmark derived from SATA-Bench. Our framework improves Macro-F1 from 48.3 to 77.0 on the human-validated \LCQA~split and from 47.6 to 75.6 on \LSATA, with the largest gains on \textsc{Neither/Nor}. %Relative calibration performs best overall, particularly on \textsc{Mixed} instances.
\end{abstract}

\section{Introduction}

%Large language models (LLMs) have significantly advanced capabilities across a wide range of natural language processing (NLP) tasks \citep{brown2020language, ouyang2022training}. By scaling up parameters and training on massive text corpora, these models excel at capturing complex linguistic patterns, retrieving factual knowledge, and generating highly fluent text \citep{kaplan2020scaling, hoffmann2022training}. Current research focuses on deploying these models to solve complex, multi-step logical reasoning problems.

Large language models (LLMs) perform well across a wide range of tasks \citep{brown2020language, ouyang2022training}, but systematic evaluations reveal persistent weaknesses in their logical reasoning \citep{parmar2024logicbench}. 
%However, systematic evaluations reveal that even strong language models struggle with complex logical reasoning and problems involving negation \citep{parmar2024logicbench}. Recent benchmarks show that models perform relatively well on conjunctive compositions, moderately well on disjunctive compositions, but collapse sharply on negation-based compositions \citep{junias-pacheco-2026-logical}. %For example, Llama-3.1-8B \citep{grattafiori2024llama} achieves only 13.1\% zero-shot macro-F1 on negation \citep{junias-pacheco-2026-logical}. 
%Negation in particular is a recurrent failure point, as models have been show to struggle revising affirmative predictions once a proposition is negated \citep{garcia2023not}, and they degrade under small shifts in wording, polarity, or scope, or when multiple negative expressions co-occur \citep{kassner2020negated, ravichander2022condaqa, she-etal-2023-scone}. %That the difficulty tracks the operator and not the content suggests that it stems from the way the logical possibilities are represented and combined. 
%However, how difficulty varies across logical operators is less understood. 
These failures are not uniform across logical operators. %Operator-level differences have been examined in several settings, but the reported orderings conflict depending on task format and how composition is evaluated~\cite{sanyal-etal-2022-robustlr,ghosh2025logical}. 
For example, \citet{junias-pacheco-2026-logical} evaluate composition over compound answer options and report a graded pattern, with performance strongest on conjunction, weaker on disjunction, and collapsing on negated compositions. That the difficulty tracks the operator rather than the content suggests it stems from the way logical possibilities are represented and combined, not from missing knowledge.

%This view is consistent with studies in human cognitive psychology, where logical reasoning imposes substantial cognitive demands even when the individual propositions are simple \citep{Klauer01021997, MEISER2001303, neys2006dual, dewall2008evidence}. If composition is a distinct burden rather than a byproduct of comprehension, then requiring a model to evaluate atomic content and combine it in a single pass conflates two operations that might be better handled separately.

\begin{figure}[!t]
    \centering
    \includegraphics[
        width=\columnwidth,
        keepaspectratio
    ]{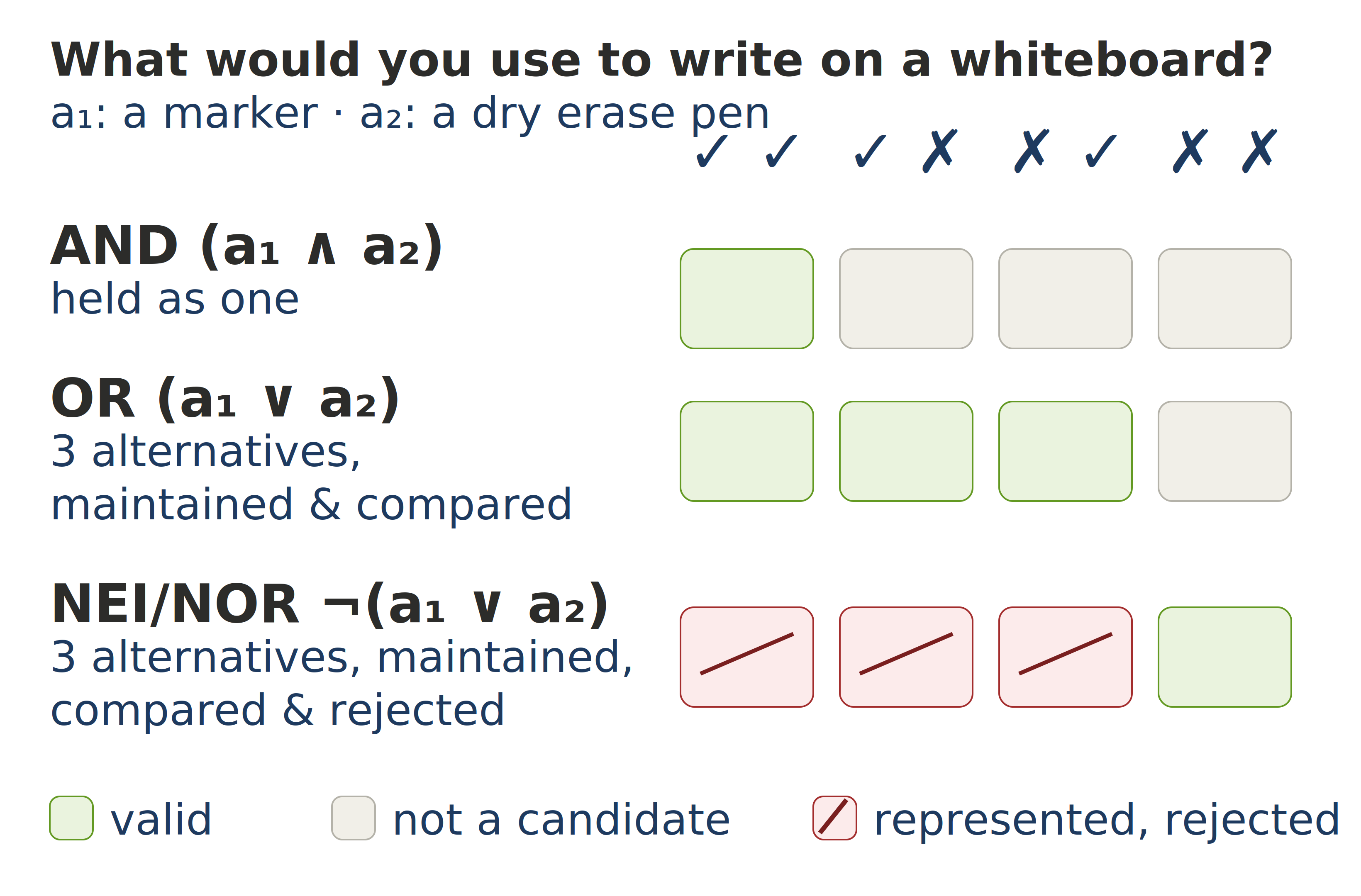}
    \caption{\small{
    Mental-model representation of the possibilities associated with
    \textsc{and}, \textsc{or}, and \textsc{neither/nor%(nei/nor)
    }.
    }}
    \label{fig:mental-model-operators}
\end{figure}

As illustrated in Figure~\ref{fig:mental-model-operators}, mental-model theories predict exactly this ordering in humans. They propose that people reason by constructing representations of situations compatible with a logical expression rather than applying formal proof rules \citep{johnson1992propositional}, so difficulty depends on the number and structure of the possibilities that must be maintained \citep{Klauer01021997, MEISER2001303, neys2006dual, dewall2008evidence}. A conjunction $(A \land B)$ can often be held as a single joint possibility, whereas a disjunction $(A \lor B)$ requires the alternatives to be maintained and compared~\citep{garcia2001conjunctive}. Negation increases difficulty, as the reasoner must represent the original proposition while tracking that it is rejected \citep{macbeth2014mental, khemlani2014negations}. \textsc{Neither/Nor} is the extreme case, combining both demands. %Psycholinguistic studies similarly show that negative sentences can take longer to represent than affirmative ones, particularly when several possible alternatives must be considered \citep{kaup2006processing, orenes2014negation}. Other theories propose that negation weakens or inhibits an activated semantic representation rather than replacing it with one clear opposite \citep{beltran2021inhibitory, zuanazzi2024negation}. These findings are especially relevant to \textsc{Neither/Nor} because the reasoner must first establish that each atomic alternative is excluded before combining the two negative judgments under the operator.
These studies do not imply that LLMs reason as humans do, but models show the same signature, degrading on disjunctive compositions \citep{khalid-etal-2025-large,10.1007/978-3-032-21289-4_7,junias-pacheco-2026-logical} and failing to revise affirmative predictions once a proposition is negated \citep{garcia2023not, kassner2020negated, ravichander2022condaqa, she-etal-2023-scone}. 

Composition is therefore a distinct burden, not a byproduct of comprehension. Standard prompting, however, fuses the two, as the model must assess each atomic proposition and combine them under the operator in a single pass. This produces a \emph{compositionality gap}, where a model solves the component subproblems correctly yet fails to combine them \citep{press-etal-2023-measuring}. %Compound answer reasoning makes this gap explicit, since the atomic answers can be evaluated independently of the rule that combines them. %This raises a critical question: can we improve compound-answer reasoning by eliciting evidence for the atomic components separately and enforcing their logical composition explicitly?
It also leaves no way to diagnose which step broke down, and it cannot enforce the composition, since a model asked to satisfy hard constraints in free generation can silently violate them. 

A common response to this gap is to make intermediate structure explicit, through chain-of-thought and decomposed prompting, entailment trees, or contrastive judgments over opposing candidates \citep{wei2022chain, khot2022decomposed, dalvi2021explaining, liusie-etal-2024-efficient}. These produce richer intermediate evidence, but the combination step remains an unconstrained generation. Neuro-symbolic approaches instead delegate inference to an external solver, first translating the natural language problem into a formal representation
%combine LLMs with external solvers by first translating the natural-language problem into its formal representation. For example, Logic-LM generates symbolic problem formulations, LINC translates premises and conclusions into first-order logic, and SatLM produces declarative specifications that are passed to an automated solver 
\citep{pan-etal-2023-logic, olausson-etal-2023-linc, ye2023satlm}. This enforces the composition, but shifts
the burden onto auto-formalization, and the solver is only as reliable as the translation it receives. In this paper, we study compound answer reasoning as a setting for isolating logical composition. This requires no translation, as the logical structure is already explicit in the answer options. What remains is to produce reliable intermediate evidence, as in the decomposition methods above, and to combine it under constraints that cannot be violated.

%These approaches delegate the main deductive process to the solver, but depend on the LLM accurately formalizing the input problem.
%Our framework differs in an important way. Rather than translating the problem into a separate formal representation, we directly use the boolean operators already present in the answer choices. The LLM evaluates whether each atomic answer is supported by the context, and the solver combines these local judgments according to the corresponding operator semantics. Our framework is closely related to prompt-based structured prediction, where local model scores are combined under global constraints, rather than translating the entire natural-language problem into a formal representation before reasoning \citep{mehta-etal-2024-promptly, pauk-pacheco-2026-mapping}.

\begin{figure*}[t]
    \centering
    \includegraphics[
        width=\textwidth,
        height=0.35\textheight,
        keepaspectratio
    ]{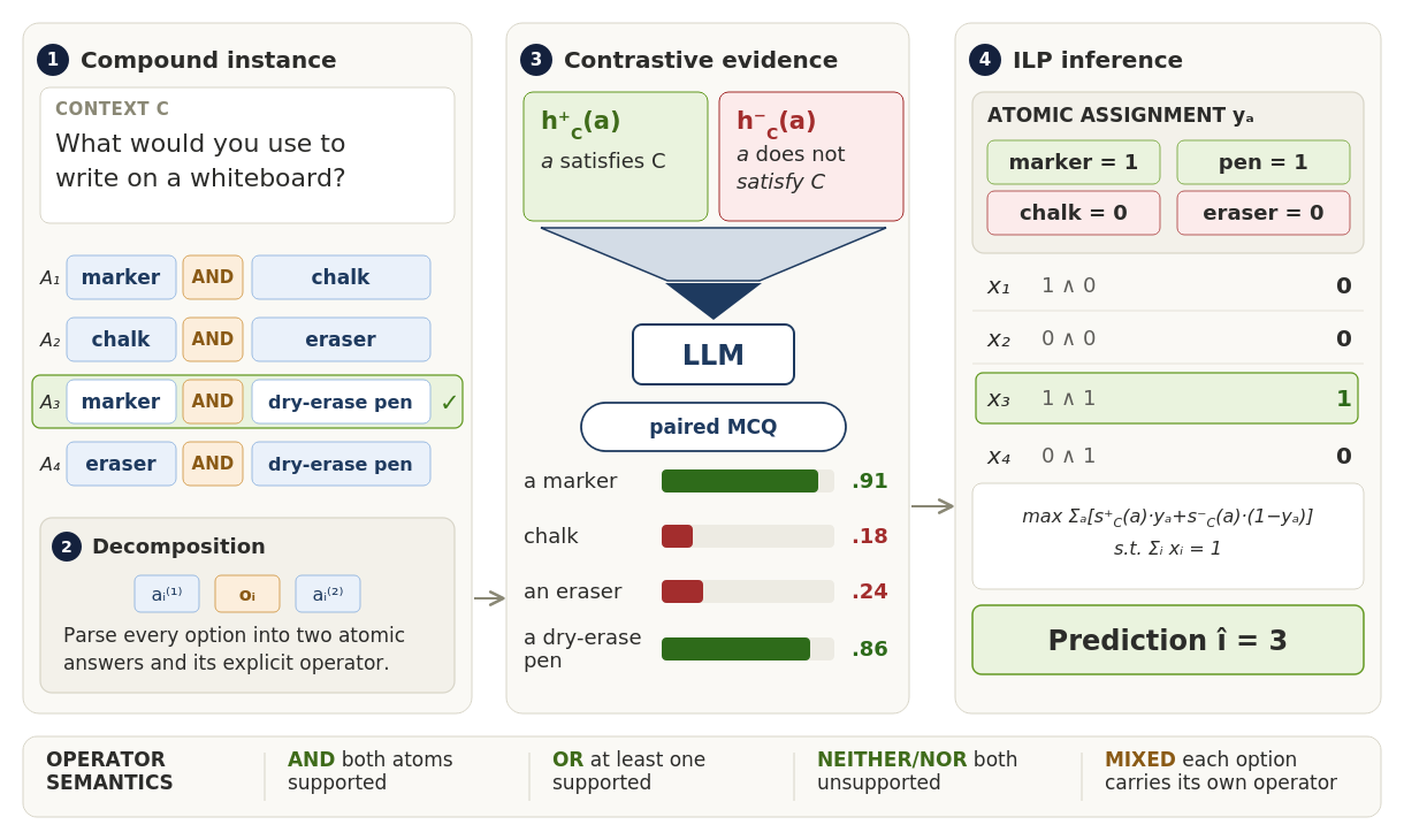}
    \caption{
    \small{Overview of the proposed framework. Each compound answer option is decomposed into two atomic answers and its explicit operator. The LLM scores local evidence for each atom, and an operator-constrained ILP combines these scores to select exactly one answer option under the corresponding operator semantics.
    }}
    \label{fig:framework}
\end{figure*}

%To address the compositionality gap, 
We present a framework that does this, decomposing each option into its atomic answers, eliciting contrastive evidence for each, and composing that evidence under the constraints imposed by the operators. %Inspired by work on human decision-making in which alternatives are evaluated through their component attributes
%\citep{tversky1972elimination}, 
%We decompose each option into its
%atomic answers and combine them under a strict composition rule. 
For every unique atomic answer, we construct paired positive and negative hypotheses stating that the answer is supported or not supported by the context. Atoms shared across options are scored once, so a proposition receives a single judgment wherever it appears. The LLM scores both hypotheses, and their contrast forms the evidence for that answer, so the model is never asked to judge a compound option as a whole. %Because these raw scores are not comparable across instances, we calibrate each one using both its value and its position among the other scores in the same instance. 
The resulting scores are calibrated and passed to an operator-constrained integer linear program (ILP), which jointly infers the assignment of each atomic answer and selects exactly one compound option. %, enforcing the operator semantics for \textsc{And}, \textsc{Or}, and \textsc{Neither/Nor}, while selecting the answer most consistent with the atomic evidence. 
Figure~\ref{fig:framework} provides an overview of the proposed framework. 
%This places our framework closer to prompt-based structured prediction, where local model scores are combined under global constraints \citep{mehta-etal-2024-promptly, pauk-pacheco-2026-mapping}, than to solver-based formalization. 

To evaluate our framework, we use two benchmarks containing compound answer options connected by \textsc{And}, \textsc{Or}, and \textsc{Neither/Nor}: \LCQA~\citep{junias-pacheco-2026-logical}, a commonsense reasoning benchmark, and \LSATA, a reading comprehension benchmark which we construct from SATA-Bench~\citep{xu2025sata}. %While \LCQA~evaluates logical composition over contextual commonsense plausibility, \LSATA~evaluates the same operator structure over paragraph-based reading-comprehension questions. 
The two require different kinds of atomic evidence, and we improve substantially on both, most sharply on the operator our account identifies as hardest: on \textsc{Neither/Nor}, macro-F1 rises from 13.9 to 76.8 on \LCQA~and from 15.2 to 73.4 on \LSATA. %\textcolor{red}{Verify}

In sum, we contribute: (1) a structured framework that elicits contrastive evidence for individual atomic answers and combines them through operator-constrained ILP inference; (2) a relative calibration method, which re-scores each atom by both its confidence and its standing among the other atoms; (3) \LSATA, a new reading-comprehension benchmark for compound answer reasoning; (4) an evaluation across two distinct benchmarks, showing the largest gains on the operators that degrade most under standard prompting.

The datasets\footnote{\scriptsize{\href{https://huggingface.co/datasets/ojayy/logical-csqa}{ojayy/logical-csqa}, \href{https://huggingface.co/datasets/ojayy/logical-sata}{ojayy/logical-sata}}}
and code\footnote{\scriptsize{\href{https://github.com/obedjunias19/structured-compositional-reasoning}{github.com/obedjunias19/structured-compositional-reasoning}}} are publicly available.

\section{Related Work}

\noindent \textbf{Logical and Compositional Reasoning with LLMs}
Logical-reasoning benchmarks such as ProofWriter \citep{tafjord-etal-2021-proofwriter}, LogicNLI \citep{tian-etal-2021-diagnosing}, FOLIO \citep{han-etal-2024-folio}, ReClor \citep{yu2020reclor}, and LogiQA \citep{ijcai2020p501} assess whether models can derive valid conclusions given facts, rules, premises, or constraints. Other datasets, including ConjNLI \citep{saha2020conjnli}, CONDAQA\citep{ravichander2022condaqa}, SCoNE \citep{she-etal-2023-scone}, and the NOT benchmark \citep{garcia2023not} isolate logical phenomena such as conjunction, disjunction and negation. These evaluations show that logical performance remains sensitive to inference structure, linguistic formulation, and negation \citep{parmar2024logicbench}.

One strategy to address this is to decompose reasoning into intermediate steps, making them explicit or dividing a complex problem into simpler subproblems \citep{wei2022chain, zhou2022least, khot2022decomposed}. EntailmentBank organizes explanations as trees of multi-premise entailment steps \citep{dalvi2021explaining}, and DecompNLI provides a systematic framework for evaluating the validity of decomposed textual inferences \citep{weir-etal-2024-enhancing}. %These methods primarily decompose questions, reasoning chains, or intermediate conclusions, but the step that combines the resulting judgments remains unconstrained. %However, models must still produce reliable component judgments and combine them correctly. \citet{press-etal-2023-measuring} describe this failure as the \emph{compositionality gap}, in which models correctly answer component questions but fail to combine those answers into the correct final prediction. In compound-answer reasoning, this structure is explicit within each option, making it possible to evaluate the atomic answers separately from the logical operator that combines them.

\paragraph{Multi- and Compound-Answer Benchmarks}
Most multiple-choice reasoning benchmarks require selecting a single correct answer type and score each candidate as a whole \citep{talmor-etal-2019-commonsenseqa, bisk2020piqa, sap-etal-2019-social, clark2018think, hendrycks2020measuring}. %Although these benchmarks evaluate different forms of knowledge and reasoning, they generally score each candidate answer as a whole. 
Consequently, they do not test whether a model can evaluate several atomic answers and combine them under an explicit logical operator.

Multi-answer question-answering benchmarks relax the assumption that each question has only one correct response. MultiRC evaluates reading comprehension questions for which several candidate answers may be correct, while RoMQA requires models to recover multiple valid answers supported by evidence distributed across passages \citep{khashabi-etal-2018-looking, zhong-etal-2023-romqa}. SATA-Bench more directly studies the select-all-that-apply format across several domains, where each option is evaluated independently and the model must identify the complete set of correct choices \citep{xu2025sata}. These benchmarks evaluate multi-answer selection, but they do not place explicit boolean operators within the candidate answers. \LCQA~does exactly this \citep{junias-pacheco-2026-logical}. We extend this operator-based structure to paragraph-based reading comprehension through \LSATA, which constructs compound options from the annotated answers in SATA-Bench. Its construction is described in Section~\ref{sec:datasets}.

\paragraph{Confidence Elicitation and Contrastive Judgments} Structured inference requires local scores that represent the model's evidence for each atomic decision. Prior work obtains such scores from token probabilities, true--false self-assessment, repeated generation, or verbalized confidence \citep{jiang-etal-2021-know, kadavath2022language, tian-etal-2023-just, pauk-pacheco-2026-mapping}. These approaches differ in whether they require access to model probabilities and in how closely their reported confidence corresponds to empirical correctness.

Other work studies comparative and contrastive judgments, where models evaluate competing candidates or opposing interpretations rather than assigning an isolated score to one statement. Such comparisons have shown advantages over pointwise evaluation in some natural language understanding and question-answering settings \citep{fortier-dubois-rosati-2023-using, liusie-etal-2024-efficient, YAO2026130407}. This line of work motivates eliciting separate evidence for positive and negative interpretations of an atomic answer before combining those judgments through structured inference.

\paragraph{Neuro-Symbolic Methods and Structured Inference} 
%With the emergence of LLMs, 
Neuro-symbolic methods increasingly use LLMs to translate NLP problems into formal representations processed by deterministic solvers \citep{pan-etal-2023-logic, ye2023satlm, olausson-etal-2023-linc}. %These methods separate language understanding from formal reasoning, but depend on the accuracy of the generated formalization.

Earlier work combines uncertain model predictions under symbolic constraints. DRaiL provides a general framework for integrating neural scorers with relational rules and global inference \citep{zhang-etal-2016-introducing, pacheco-goldwasser-2021-modeling}. Particularly relevant to question answering, \citet{pujari-goldwasser-2019-using} combine per-option machine-comprehension scores with NLI-based relations between answer choices and use ILP inference to obtain consistent predictions. Other approaches use satisfiability-based inference to reconcile model beliefs, answer compatibility relations, or generated explanations \citep{kassner-etal-2021-beliefbank,
mitchell-etal-2022-enhancing, jung-etal-2022-maieutic}.

More recent work combines prompted local predictions with combinatorial inference and studies confidence elicitation, calibration, and structured learning in this setting \citep{mehta-etal-2024-promptly, pauk-pacheco-2026-mapping}. Our framework is closer to this line of work than to full-problem autoformalization: the context and atomic answers remain in natural language, while the explicitly provided boolean operators determine how the local evidence is composed.

\section{Framework Overview}

In this section, we present a framework for compound answer reasoning in which an LLM supplies local atomic evidence and a structured inference layer performs the logical composition. We parse each option into its atomic answers and operator (Sec. \ref{subsec:option_decomposition}), elicit and calibrate evidence for each atomic answer in isolation (Sec. \ref{subsec:hypothesis_construction}--Sec. \ref{subsec:score_calibration}), and defer composition to an integer linear program that combines this evidence under the operator semantics (Sec. \ref{subsec:ilp_inference}) Fig. \ref{fig:framework} gives an overview.

\subsection{Task Formulation}
\label{subsec:task-formulation}
Each instance consists of a context $C = (P, q)$, where $q$ is a question and $P$ is an optional paragraph needed to answer it, together with four candidate answer options $\mathcal{A} = \{A_1, A_2, A_3, A_4\}$.
%For \LCQA, $C$ is a commonsense question $q$, whereas for \LSATA, $C = (P, q)$ consists of a paragraph $P$ and its corresponding reading-comprehension question $q$.
Unlike standard multiple choice, each option is \emph{compound}. That is, it contains two atomic answers joined by an explicit boolean operator,

\[\small
\begin{gathered}
A_i = a_i^{(1)} \circ_i\; a_i^{(2)}, \\[4pt]
\circ_i \in \{\textsc{And}, \textsc{Or}, \textsc{Neither/Nor}\}.
\end{gathered}
\]

For the question \emph{What would you use to write on a whiteboard?}, one option might be \emph{a marker} \textsc{And} \emph{chalk} and another \emph{a marker} \textsc{And} \emph{a dry-erase pen}. Both contain the atomic answer \emph{a marker}, but only the second is valid. Validity therefore depends on two separable things: whether each atomic answer holds in the context, and how the operator combines them. Let
$\phi_\circ : \{0,1\}^2 \rightarrow \{0,1\}$ denote the composition rule for operator $\circ$, applied to the statuses of two atomic answers:

\[\scriptsize
\phi_\circ(\rho_1,\rho_2)
=
\begin{cases}
\rho_1 \land \rho_2, & \circ = \textsc{And} \\[2pt]
\rho_1 \lor \rho_2, & \circ = \textsc{Or} \\[2pt]
\neg\rho_1 \land \neg\rho_2, & \circ = \textsc{Neither/Nor}.
\end{cases}
\]

That is, \textsc{And} requires both atoms to hold, \textsc{Or} requires at least one, and \textsc{Neither/Nor} requires that neither does. In the example, \emph{a marker} and \emph{a dry-erase pen} both hold while \emph{chalk} does not, so $\phi_{\textsc{And}}$ returns $1 \land 0 = 0$ for the first option and $1 \land 1 = 1$ for the second.

%For a given context $C$, let $\rho_C(a) \in \{0,1\}$ denote the ground-truth binary label of an atomic answer $a$. %In \LCQA, we set $\rho_C(a) = 1$ when $a$ is a contextually plausible answer to the commonsense question, and $\rho_C(a) = 0$ otherwise. In \LSATA, we set $\rho_C(a) = 1$ when $a$ is a correct answer to the reading-comprehension question, and $\rho_C(a) = 0$ otherwise. 
%In other words, $\rho_C(a) = 1$ represents the case where an atomic answer satisfies the given context, and $\rho_C(a) = 0$ the case where it does not. %Here, $\rho_C(a)$ denotes the ground-truth binary label rather than the prediction produced by a particular model.

Every instance is constructed so that exactly one option is valid under the gold statuses of its atoms. This is what makes the task diagnostic, as a model cannot succeed by scoring options independently, since the correct answer is
determined jointly by the atomic statuses and the operators applied to them. The task is to predict the index $i^*$ of that option. Written this way, the atomic statuses carry the content of the task and $\phi_\circ$ carries its logic. %Our framework mirrors this split. The model supplies evidence about the atoms, while $\phi_\circ$ is enforced exactly by the inference procedure of \S\ref{subsec:inference}, which recovers a predicted status $y_a$ for each atom and a predicted validity $x_i$ for each option.

%Both benchmarks contain three operator-specific settings, in which all four options use the same operator, and a \textsc{Mixed} setting, in which operators may vary across options.

\subsection{Option Decomposition}
\label{subsec:option_decomposition}

We deterministically parse each option into the triplet $A_i = (a_i^{(1)}, \circ_i, a_i^{(2)})$ % we parse the option text to extract its two constituent atomic answers, $a_i^{(1)}$ and $a_i^{(2)}$, along with the operator $o_i$ that connects them. 
and collect the atomic answers appearing
anywhere in the instance:

\[\scriptsize{
\mathcal{U}_C
=
\bigcup_{i=1}^{4}
\left\{
a_i^{(1)}, a_i^{(2)}
\right\}.}
\]

Because $\mathcal{U}_C$ collects atomic answers rather than options, an atom occurring in several options appears in it once. In our running example, \emph{a marker} contributes a single element even though it occurs in two options, and any status later assigned to it applies to both. %The elements of $\mathcal{U}_C$, not the compound options, are the units the rest of this section operates on.
%Rather than asking the model to directly evaluate each compound option, every atomic answer $a \in \mathcal{U}_C$ is evaluated separately with respect to the shared context $C$.

\subsection{Contrastive Hypothesis Construction}
\label{subsec:hypothesis_construction}

To supply evidence about the status of an atomic answer $a \in \mathcal{U}_C$, we construct a pair of opposing hypotheses conditioned on $C$: %Following prior work on comparative and contrastive evaluation \citep{liusie-etal-2024-efficient}, we represent the two possible task-level interpretations of $a$ as

\[\small
\begin{aligned}
h_C^{+}(a) &: \quad a \text{ satisfies context } C, \\[4pt]
h_C^{-}(a) &: \quad a \text{ does not satisfy context } C.
\end{aligned}
\]

\noindent For the whiteboard example, the pair for \emph{chalk} asserts that chalk is, and is not, something you would use to write on a whiteboard.

Eliciting evidence for both members of the pair makes the result a comparison between two readings of the same atom. Prior work on comparative and contrastive evaluation finds such judgments more reliable \citep{liusie-etal-2024-efficient}.
%In \LCQA, the hypotheses express whether the atomic answer is contextually plausible, whereas in \LSATA, they express whether the atomic answer is correct given the paragraph and reading-comprehension question.
%At this stage, we do not determine which element in the pair is correct. We only produce the opposing statements that will be evaluated in the next steps.

\subsection{Confidence Elicitation}
\label{subsec:confidence_elicitation}

Next, we estimate the model's local evidence for the two hypotheses associated with each atomic answer. Following prior work on prompt-based structured prediction \citep{pauk-pacheco-2026-mapping}, we present $h_C^{+}(a)$ and
$h_C^{-}(a)$ as choices \texttt{A} and \texttt{B} within a single prompt and ask the model which is more plausible. Let $\ell_C^{+}(a)$ and $\ell_C^{-}(a)$ denote the log probabilities of the first answer tokens corresponding to the
two choices. The raw evidence scores are their normalization over the two alternatives:

%we investigate white-box and black-box strategies that produce non-negative support values
%\[
%r_C{+}(a),\,
%r_C{-}(a).
%\]
%For all methods, we convert these values into relative evidence scores using
%\begin{align}
%s_{C,\mathrm{raw}}^{+}(a)
%&=
%\frac{r_C^{+}(a)}
%{r_C^{+}(a)+r_C^{-}(a)},
%\\[6pt]
%s_{C,\mathrm{raw}}^{-}(a)
%&=
%\frac{r_C^{-}(a)}
%{r_C^{+}(a)+r_C^{-}(a)}.
%\label{eq:relative-evidence-normalization}
%\end{align}

\begin{equation}\scriptsize{
s_{C,\mathrm{raw}}^{\pm}(a)
=
\frac{\exp\!\left(\ell_C^{\pm}(a)\right)}
{\exp\!\left(\ell_C^{+}(a)\right)+\exp\!\left(\ell_C^{-}(a)\right)}
}\label{eq:paired-mc}
\end{equation}

%The resulting scores lie in $[0,1]$, sum to one and quantify the model's local evidence over the two opposing hypotheses.

The two scores lie in $[0,1]$ and sum to one, so either determines the other. %Because both hypotheses appear in the same prompt, the resulting score is a comparison between two readings of $a$ rather than an isolated judgment about one of them. 
These are the model's local evidence for $a$, obtained before any logical constraint is applied.

We also evaluated three alternatives, which differ only in how support for the two hypotheses is obtained: independent true--false scoring, generation sampling, and verbalized confidence. Paired multiple choice provides the strongest evidence on our main experiments (Sec. \ref{sec:main-results}). We describe the alternatives in App.~\ref{app:elicitation}. 
% \textcolor{red}{Move that content to appendix and reference it here}

\subsection{Score Calibration}
\label{subsec:score_calibration}

Equation~\ref{eq:paired-mc} gives the model's relative preference between the two hypotheses, not the probability that $a$ is correct. Since these scores are later combined across the atomic answers of an instance, they need to be comparable to one another, and a raw preference of $0.9$ need not carry the same weight for one atom as for another. We therefore calibrate them against the gold atomic status. Because the positive and negative scores sum to one, calibrating the positive score determines the negative one, $s_{C,\mathrm{cal}}^{-}(a) = 1 - s_{C,\mathrm{cal}}^{+}(a)$. All calibrators are fit on atomic examples from the training set.

%The confidence-elicitation methods produce continuous scores representing the model's relative preference for the positive and negative hypotheses associated with each atomic answer. These scores, however, do not necessarily represent the probability that the atomic answer is correct. This distinction is important because the ILP compares evidence across multiple atomic answers when selecting a globally consistent assignment. We therefore calibrate the elicited scores before using them in the ILP objective.

%For an atomic answer $a$, let $s_{C,\mathrm{raw}}^{+}(a)$ denote the uncalibrated evidence score for its positive hypothesis. Because the positive and negative evidence scores sum to one, calibrating the positive score also determines the negative score. We therefore calibrate only \(s_{C,\mathrm{raw}}^{+}(a)\) and define
%\[
%s_{C,\mathrm{cal}}^{-}(a)
%=
%1-s_{C,\mathrm{cal}}^{+}(a).
%\]

%The calibrators are fitted using atomic examples from the training set. We evaluate two standard post-hoc methods, Platt scaling and isotonic calibration, together with our proposed relative calibration method.

We evaluate two standard post-hoc calibration methods. Platt scaling fits a logistic transformation of the raw positive score to the gold atomic label \citep{platt1999probabilistic}. Isotonic calibration instead fits a non-decreasing non-parametric mapping, without assuming a sigmoid relationship between the score and correctness \citep{zadrozny2002transforming}.

%\subsubsection{Platt Scaling}
%Platt scaling fits a logistic transformation from the raw positive evidence score to the gold atomic label \citep{platt1999probabilistic}. The learned mapping produces the calibrated positive score, while the calibrated negative score is obtained as its complement.

%\subsubsection{Isotonic Calibration}
%Isotonic calibration learns a non-decreasing, non-parametric mapping from the raw positive evidence score to the gold atomic label \citep{zadrozny2002transforming}. Unlike Platt scaling, it does not assume a sigmoid relationship between the raw score and atomic correctness.

\paragraph{Relative Calibration}
Platt scaling and isotonic calibration adjust each atomic score based only on its absolute value. However, exactly one option in an instance is valid, so which atoms are selected depends on how they stand relative to the others present. Two atoms scored $0.91$ and $0.86$ support different conclusions depending on whether the remaining atoms sit near $0.9$ or near $0.2$, and an independent mapping cannot distinguish these cases. We therefore introduce \textit{relative calibration}, a supervised re-scoring method that supplies both a score's magnitude and its standing within the instance.
%However, the ILP compares evidence across the atomic answers within an instance. We therefore introduce relative calibration, which considers both the raw positive score and its position relative to the other atomic scores in the same instance.

For each atomic answer $a$, we construct the feature vector

{\scriptsize
\[
\mathbf{f}_C(a)
=
\begin{bmatrix}
\operatorname{logit}\!\left(s_{C,\mathrm{raw}}^{+}(a)\right) \\
z_C(a) \\
\operatorname{rank}_C(a) \\
s_{C,\mathrm{raw}}^{+}(a) -  s_{C,\max}^{+}
\end{bmatrix}.
\]
}
where $z_C(a)$ is the within-instance standardized score, $\operatorname{rank}_C(a)$ is its rank among the atomic scores, and $s_{C,\max}^{+}$ is the highest positive score in the instance. A logistic regression maps these features to the calibrated positive score:
{\scriptsize\[
s_{C,\mathrm{cal}}^{+}(a)
=
\sigma\!\left(\mathbf{w}^{\top}\textbf{f}_C(a) + b\right),
\]}
with $\mathbf{w}$ and $b$ learned from the training set. %Relative calibration is suited to our framework because global inference compares alternative atomic assignments within each instance rather than evaluating each atomic score in isolation.

\subsection{Globally Constrained Inference}
\label{subsec:ilp_inference}

The confidence elicitation and calibration stages produce continuous local evidence for each atomic answer, with no reference to the operators. Following prompt-based structured prediction \citep{mehta-etal-2024-promptly,pauk-pacheco-2026-mapping}, we combine these scores under global constraints that encode the logical structure of the compound answer options using ILP. Going forward, we write $s_{C}^{\pm}(a)$ for the calibrated scores $s_{C,\mathrm{cal}}^{\pm}(a)$. The uncalibrated variant we report in Sec. \ref{sec:experiments} substitutes $s_{C,\mathrm{raw}}^{\pm}(a)$ throughout. %Let $e_C^{+}(a)$ and $e_C^{-}(a)$ denote the positive and negative evidence supplied to the inference procedure for atomic answer $a$. These values depend on whether calibration is applied:
%\[
%e_C^{\pm}(a)
%=
%\begin{cases}
%s_{C,\mathrm{raw}}^{\pm}(a),
%& \text{without calibration}, \\[4pt]
%s_{C,\mathrm{cal}}^{\pm}(a),
%& \text{with calibration}.
%\end{cases}
%\]
%Here, $s_{C,\mathrm{raw}}^{\pm}(a)$ denotes the evidence obtained directly from the confidence-elicitation strategy, whereas $s_{C,\mathrm{cal}}^{\pm}(a)$ denotes the corresponding evidence after supervised calibration.

%Let $\mathcal{U}_C$ denote the set of unique atomic answers appearing across the four options for context $C$.

%We formulate a binary Integer Linear Program (ILP) that finds a single support assignment over all unique atomic answers in $\mathcal{U}_C$. The validity of each compound option is then determined from this shared assignment according to its associated Boolean operator. The ILP maximizes the total local evidence while requiring exactly one compound option to be valid.

\paragraph{Decision Variables}
We formulate a binary ILP over two sets of variables. For each atomic answer $a \in \mathcal{U}_C$, $y_a \in \{0,1\}$ is its inferred
status, with $y_a = 1$ when $a$ satisfies $C$. Similarly, for each compound option $A_i = a_i^{(1)} \circ_i\; a_i^{(2)}$, $x_i \in \{0,1\}$ is its inferred validity, taking the value of 1 when the statuses assigned to its two atomic answers satisfy the operator $\circ_i$ and 0 otherwise. %Because $\mathcal{U}_C$ holds each atomic answer once, an atom occurring in several options has a single variable, and the status it receives applies to all of them. 
%We define the following binary optimization variables:
%\begin{itemize}
%\item $y_a \in \{0,1\}$ represents the inferred binary assignment for atomic answer $a \in \mathcal{U}_C$, where $y_a = 1$ denotes that $a$ satisfies context $C$, and $y_a = 0$ denotes that it does not.
%\item $x_i \in \{0,1\}$ represents the logical validity of compound option $A_i$, where $i \in \{1,2,3,4\}$. Specifically, $x_i = 1$ when the binary assignments of its two atomic answers satisfy the Boolean operator associated with $A_i$, and $x_i = 0$ otherwise.
%\end{itemize}
%Given that our task formulation requires exactly one compound option to be valid, the unique option for which $x_i = 1$ is returned as the final prediction. 
We collect all these variables into $\mathbf{y} \in \{0,1\}^{|\mathcal{U}_C|}$ and $\mathbf{x} \in \{0,1\}^{4}$.

\paragraph{Operator Constraints}
The validity of each compound option $A_i = a_i^{(1)} \circ_i\; a_i^{(2)}$ must equal the composition rule of Sec. \ref{subsec:task-formulation} applied to the inferred statuses of its atoms,
\[\scriptsize{
x_i = \phi_{\circ_i}\!\left(y_{a_i^{(1)}},\, y_{a_i^{(2)}}\right),}
\]

%Let $a_i^{(1)}$ and $a_i^{(2)}$ denote the two atomic answers contained in option $A_i$. 
%We constrain $x_i$ to be exactly equal to the Boolean composition of the inferred binary assignments of its two atomic answers.
\noindent which we encode exactly with linear inequalities. Writing $y_1$ and $y_2$ for $y_{a_i^{(1)}}$ and $y_{a_i^{(2)}}$, respectively:

\[\scriptsize{
\begin{array}{ll}
\textsc{And}: & x_i \leq y_1,\ \ x_i \leq y_2, \\
 & x_i \geq y_1 + y_2 - 1 \\[5pt]
\textsc{Or}: & x_i \geq y_1,\ \ x_i \geq y_2, \\
 & x_i \leq y_1 + y_2 \\[5pt]
\textsc{NNor}: & x_i \leq 1 - y_1,\ \ x_i \leq 1 - y_2, \\
 & x_i \geq 1 - y_1 - y_2
\end{array}}
\]

%For an \textsc{And} option, both atomic answers must be supported:
%\[
%x_i \leq y_{a_i^{(1)}},
%\qquad
%x_i \leq y_{a_i^{(2)}},
%\]
%\[
%x_i
%\geq
%y_{a_i^{(1)}} + y_{a_i^{(2)}} - 1.
%\]
%For an \textsc{Or} option, at least one atomic answer must be supported:
%\[
%x_i \geq y_{a_i^{(1)}},
%\qquad
%x_i \geq y_{a_i^{(2)}},
%\]
%\[
%x_i
%\leq
%y_{a_i^{(1)}} + y_{a_i^{(2)}}.
%\]
%For a \textsc{Neither/Nor} option, both atomic answers must be unsupported:
%\[
%x_i \leq 1 - y_{a_i^{(1)}},
%\qquad
%x_i \leq 1 - y_{a_i^{(2)}},
%\]
%\[
%x_i
%\geq
%1 - y_{a_i^{(1)}} - y_{a_i^{(2)}}.
%\]

The \textsc{Neither/Nor} constraints are the complement of the \textsc{Or} constraints, as $\neg(y_1 \lor y_2)$ requires that every assignment satisfying the disjunction be excluded. Since for our task exactly one option is valid by construction (Sec. \ref{subsec:task-formulation}), we additionally require
$\sum_{i=1}^{4} x_i = 1$.

%Together, these constraints ensure that $x_i = 1$ if and only if the inferred binary assignments of the two atomic answers satisfy the operator associated with $A_i$. In the \textsc{Mixed} setting, the corresponding constraint set is applied separately to each option according to its operator.
%Because each benchmark instance contains exactly one valid compound answer, we additionally require
%\[
%\sum_{i=1}^{4} x_i = 1.
%\]
%The ILP must therefore find an atomic assignment under which exactly one of the four candidate options is logically valid.

\paragraph{Objective}
Among the assignments satisfying these constraints, we select the one carrying the strongest total evidence by optimizing:
%The ILP selects the globally consistent atomic assignment that receives the strongest total local evidence. We optimize
\[\scriptsize{
\begin{aligned}
\max_{\mathbf{y},\, \mathbf{x}}
\quad & \sum_{a \in \mathcal{U}_C}
\left[
s_{C}^{+}(a)\, y_a
+
s_{C}^{-}(a)\,(1 - y_a)
\right] \\[3pt]
\text{s.t.} \quad & (\mathbf{y}, \mathbf{x}) \in \mathcal{F}_C
\end{aligned}}
\]

\noindent where $\mathcal{F}_C$ denotes the set of feasible assignments, that is, the pairs $(\mathbf{y}, \mathbf{x}) \in \{0,1\}^{|\mathcal{U}_C|} \times \{0,1\}^{4}$ that satisfy all constraints.

%\[
%\max_{\{x_i\}_{i=1}^{4},\,\{y_a\}_{a \in \mathcal{U}_C}}
%\sum_{a \in \mathcal{U}_C}
%\left[
%s_C^{+}(a)\, y_a
%+
%s_C^{-}(a)\,(1 - y_a)
%\right]
%\]
%For each atomic answer $a$, the first term $s_{C}^{+}(a)\, y_a$ contributes the positive evidence when $y_a = 1$, while the second $s_{C}^{-}(a)\,(1 - y_a)$ contributes the negative evidence when $y_a = 0$. %Each unique atomic answer contributes exactly once, and a shared variable $y_a$ ensures that repeated occurrences receive the same binary assignment across all options.
%The objective then selects the highest-scoring atomic assignment that satisfies all constraints. The predicted answer is the unique option index $\widehat{i}$ for which
%\[
%x_{\widehat{i}} = 1.
%\]

This objective scores only the atomic assignment, where the first term ($s_{C}^{+}$) contributes the supporting evidence, while the second ($s_{C}^{-}$) contributes the opposing evidence (Sec. \ref{subsec:hypothesis_construction}). The constraints then determine which option follows from it. The predicted answer is the unique index $\hat{\imath}$ with $x_{\hat{\imath}} = 1$.
\section{Benchmark Datasets}
\label{sec:datasets}

We evaluate our framework on two benchmarks that follow the compound answer reasoning task formulation in Section~\ref{subsec:task-formulation} but differ in the source of their atomic labels: contextual commonsense plausibility and reading-comprehension ratings.

\paragraph{\LCQA} evaluates the interaction between commonsense judgment and logical composition \citep{junias-pacheco-2026-logical}. Each instance contains a commonsense question and four compound answer options, each joining two atomic answers under a logical operator. For \emph{Where do you see tiny bottles of shampoo when away from home?}, one option is \emph{hotels} \textsc{And} \emph{gym showers}. %Figure~\ref{fig:lcsqa_structure} in the Appendix illustrates the task.
It contains 19,996 instances (11,996 train / 6,000 dev / 2,000 test), evenly distributed across four settings: three in which all four options share the same operator (\textsc{And}, \textsc{Or}, \textsc{Neither/Nor}), and a \textsc{Mixed} setting in which operators may differ across the options of an instance. The test set is further divided into human-validated (HV) and non-validated (NV) subsets of 1,000 each.

%Because a model may correctly judge which individual atomic answers are plausible yet still fail to combine them, the benchmark separates atomic commonsense knowledge from the ability to compose it under the specified operator.
%\LCQA~is useful for separating atomic commonsense knowledge from compound logical reasoning performance. A model may correctly determine which individual atomic answers are plausible but still fail to combine those judgments into the correct compound answer. The benchmark therefore allows atomic commonsense plausibility to be evaluated separately from the model's ability to compose those judgments under the specified operator.

\paragraph{\LSATA} is constructed from the human-labeled training partition of \textsc{SATA-Bench} \citep{xu2025sata}, which uses a select-all-that-apply format in which multiple answer choices may be correct for a paragraph-based reading-comprehension question. %We transform these annotations into compound multiple-choice instances.
We pair these annotated answers into compound options of the same form as \LCQA, two atomic answers joined by a logical operator. We first remove duplicate source instances and retain questions containing at least two correct and at least three incorrect answers, which is the minimum needed to build one valid option and three distractors. This yields 1,390 eligible source questions, of which 1,350 are selected to obtain balanced training, development, and test splits.

For each source question, we partition the original choices into correct and incorrect atomic answers and construct valid and invalid compound option pools according to operator semantics. %Namely, we generate compound answers for each of the \textsc{And}, \textsc{Or}, \textsc{Neither/Nor} settings. 
We construct operator-specific instances (\textsc{And}, \textsc{Or}, \textsc{Neither/Nor}), in which the gold option and distractors are sampled from the corresponding pools. We also construct \textsc{Mixed} instances, in which candidates from all three operators are combined into a single pool before sampling one valid option and three distractors. The resulting dataset contains 5,400 instances (2,400 train / 1,000 dev / 2,000 test), evenly distributed across the operator settings. 

To assess \LSATA's quality, one author and one independent annotator answered 40 instances (10 per setting) without access to the gold labels. They achieved 87.5\% and 80.0\% accuracy, respectively, with 75.0\% agreement ($\kappa=0.67$).

Only 4.30\% of \LCQA-HV and 5.15\% of \LSATA~instances can be answered from the logical form of the options alone, so most instances still require judging the atomic answers (App.~\ref{app:structural-shortcuts}). App.~\ref{app:benchmark} includes additional details and examples for both benchmarks, as well as a schema summarizing the construction of \LSATA.

\section{Experiments} \label{sec:experiments}

\paragraph{Experimental Settings}
%\label{sec:experimental-setup}
%We evaluate our framework on the \LCQA~and \LSATA~benchmarks. \LCQA~contains 19,996 instances divided into 11,996 training, 6,000 development, and 2,000 test examples. Each split is evenly distributed across \textsc{and}, \textsc{or}, \textsc{neither}, and Mixed instances. The test set is further divided into human-validated (HV) and non-validated (NV) subsets of 1,000 instances each. We report the main results on both subsets and conduct the primary calibration and error analyses on the HV subset. \LSATA~contains 5,400 instances divided into 2,400 training, 1,000 development, and 2,000 test examples. Each split is evenly distributed across \textsc{and}, \textsc{or}, \textsc{neither}, and Mixed instances. We use the full training set to fit the calibration methods and report evaluation results on the test set.
Our main experiments use \texttt{Llama-3.1-8B-Instruct} \citep{grattafiori2024llama}, hereafter \textsc{Llama}-8B, and temperature 0.7, with results averaged over five runs. We also evaluate \textsc{Llama-3.2-1B-Instruct}, \textsc{Llama-3.2-3B-Instruct}, \textsc{Llama-3.3-70B-Instruct} (4-bit AWQ), and \textsc{Ministral-8B-Instruct-2410} to test whether the results hold across model families and scales. We report Macro-F1, and Brier score and log loss for atomic calibration quality. Implementation details and prompt templates are in App.~\ref{app:implementation-details} and App.~\ref{app:prompts}.

\begin{table}[t]
\centering
\scriptsize
\setlength{\tabcolsep}{2.5pt}
\renewcommand{\arraystretch}{1.05}
\resizebox{\columnwidth}{!}{%
\begin{tabular}{lccccc}
\toprule
\textbf{Method}
& \textbf{\textsc{And}}
& \textbf{\textsc{Or}}
& \textbf{\textsc{NN}}
& \textbf{\textsc{Mix}}
& \textbf{All} \\
\midrule
\multicolumn{6}{l}{\textit{Direct prompting}} \\
0-shot
& $68.6^{\pm0.7}$
& $54.4^{\pm2.4}$
& $12.1^{\pm1.3}$
& $46.3^{\pm1.6}$
& $45.7^{\pm1.0}$ \\
%1-shot
%& $72.1^{\pm1.0}$
%& $56.6^{\pm1.7}$
%& $6.6^{\pm1.1}$
%& $44.7^{\pm1.7}$
%& $45.2^{\pm1.1}$ \\
%2-shot
%& $74.9^{\pm1.7}$
%& $60.2^{\pm2.3}$
%& $7.7^{\pm2.1}$
%& $43.9^{\pm2.4}$
%& $47.0^{\pm0.9}$ \\
3-shot
& $74.3^{\pm1.3}$
& $61.8^{\pm1.3}$
& $7.5^{\pm1.6}$
& $44.5^{\pm2.1}$
& $47.3^{\pm0.6}$ \\
CoT (0-shot)
& $70.1^{\pm1.7}$
& $62.3^{\pm0.7}$
& $13.9^{\pm0.7}$
& $46.1^{\pm2.7}$
& $48.3^{\pm0.5}$ \\
\midrule
\multicolumn{6}{l}{\textit{Structured inference}} \\
Paired MC
& $72.4^{\pm0.0}$
& $\mathbf{85.2^{\pm0.0}}$
& $75.1^{\pm0.0}$
& $70.3^{\pm0.0}$
& $75.8^{\pm0.0}$ \\
\quad + Platt
& $72.0^{\pm0.0}$
& $84.4^{\pm0.0}$
& $75.6^{\pm0.0}$
& $73.7^{\pm0.2}$
& $76.4^{\pm0.1}$ \\
\quad + Isotonic
& $72.6^{\pm0.3}$
& $83.6^{\pm0.2}$
& $75.2^{\pm0.0}$
& $71.5^{\pm0.4}$
& $75.7^{\pm0.2}$ \\
\quad + Relative
& $71.8^{\pm0.2}$
& $84.0^{\pm0.0}$
& $\mathbf{76.8^{\pm0.0}}$
& $\mathbf{75.2^{\pm0.0}}$
& $\mathbf{77.0^{\pm0.1}}$ \\
\bottomrule
\end{tabular}%
}
\caption{\small{Macro-F1 on the human-validated (HV) split of \LCQA.
NN = \textsc{Neither/Nor}.}}
\label{tab:lcsqa-main}
\end{table}

\begin{table}[t]
\centering
\scriptsize
\setlength{\tabcolsep}{2.5pt}
\renewcommand{\arraystretch}{1.05}
\resizebox{\columnwidth}{!}{%
\begin{tabular}{lccccc}
\toprule
\textbf{Method}
& \textbf{AND}
& \textbf{OR}
& \textbf{NN}
& \textbf{MIX}
& \textbf{All} \\
\midrule
\multicolumn{6}{l}{\textit{Direct prompting}} \\
0-shot
& $69.0^{\pm1.1}$
& $60.5^{\pm1.6}$
& $15.2^{\pm2.1}$
& $38.5^{\pm1.1}$
& $46.2^{\pm0.9}$ \\
%1-shot
%& $71.0^{\pm1.0}$
%& $60.3^{\pm0.9}$
%& $11.2^{\pm1.0}$
%& $35.5^{\pm1.2}$
%& $44.6^{\pm0.4}$ \\
%2-shot
%& $69.7^{\pm0.8}$
%& $59.5^{\pm1.5}$
%& $10.1^{\pm1.6}$
%& $37.8^{\pm1.2}$
%& $44.4^{\pm0.7}$ \\
3-shot
& $68.4^{\pm0.9}$
& $59.1^{\pm0.9}$
& $11.0^{\pm1.9}$
& $36.2^{\pm1.6}$
& $43.9^{\pm1.1}$ \\
CoT (0-shot)
& $70.5^{\pm1.7}$
& $65.2^{\pm1.4}$
& $14.5^{\pm2.4}$
& $39.2^{\pm2.6}$
& $47.6^{\pm1.4}$ \\
\midrule
\multicolumn{6}{l}{\textit{Structured inference}} \\
Paired MC
& $73.6^{\pm0.0}$
& $82.2^{\pm0.0}$
& $71.9^{\pm0.0}$
& $60.9^{\pm0.0}$
& $72.2^{\pm0.0}$ \\
\quad + Platt
& $\mathbf{74.8^{\pm0.0}}$
& $82.4^{\pm0.0}$
& $73.0^{\pm0.0}$
& $70.4^{\pm0.0}$
& $75.2^{\pm0.0}$ \\
\quad + Isotonic
& $74.6^{\pm0.0}$
& $\mathbf{83.7^{\pm0.0}}$
& $73.0^{\pm0.0}$
& $69.7^{\pm0.0}$
& $75.3^{\pm0.0}$ \\
\quad + Relative
& $74.4^{\pm0.0}$
& $82.2^{\pm0.0}$
& $\mathbf{73.4^{\pm0.0}}$
& $\mathbf{72.1^{\pm0.0}}$
& $\mathbf{75.6^{\pm0.0}}$ \\
\bottomrule
\end{tabular}%
}
\caption{\small{Macro-F1 on the \LSATA~test set.}}
\label{tab:lsata-main}
\end{table}

\subsection{Main Results}
\label{sec:main-results}

%Tables~\ref{tab:lcsqa-main} and~\ref{tab:lsata-main} compare our structured-inference framework with LLaMA-8B under direct 0--3-shot and 0-shot chain-of-thought settings. We report Macro-F1 on the human-validated (HV) split of \LCQA~and the \LSATA~test set as the mean $\pm$ standard deviation over five runs. Direct prompting uses first-token log-probabilities over options A-D. We also evaluate generation-based direct prompting which achieves slightly lower overall Macro-F1 on both benchmarks (App. ~\ref{app:generation-direct}). Paired multiple-choice scoring outperforms alternative confidence elicitation strategies overall on both benchmarks (App~\ref{app:elicitation-results}). 

Tables~\ref{tab:lcsqa-main} and~\ref{tab:lsata-main} compare our framework with direct prompting of \textsc{Llama-8B} (0-shot, 3-shot, and 0-shot CoT) on \LCQA-HV and \LSATA. Direct prompting uses first-token log-probabilities over options A--D, as a generation-based variant performs slightly worse (App.~\ref{app:generation-direct}). Paired multiple choice provides the strongest evidence for our main model (App.~\ref{app:elicitation-results}). %and we describe the alternatives in App.~\ref{app:elicitation}.

\paragraph{Direct Prompting vs. Structured Inference}
Structured inference substantially outperforms direct prompting across both benchmarks. On the \LCQA-HV split, macro-F1 for the strongest direct-prompting configuration is 48.3, whereas our paired multiple-choice evidence with globally constrained inference achieves 75.8, an improvement of 27.5 points. Relative calibration further increases the performance to 77.0. The same pattern holds for \LSATA~benchmark, where paired multiple-choice structured inference obtains 72.2 macro-F1, compared to 47.6 for the strongest direct-prompting baseline. These results provide evidence that explicitly separating atomic evaluation from logical composition can reduce the compositionality gap observed under direct compound-answer prediction. %\new{Results across four additional models show that structured inference outperforms direct prompting (App~\ref{app:other-models}).} 

\paragraph{Generalization Across Models}
The same pattern holds for four additional models from 1B to 70B parameters (App.~\ref{app:other-models}). Structured inference with paired MC improves overall Macro-F1 over the best direct-prompting baseline for every model, by 17.8 to 31.5 points on \LCQA-HV and 19.4 to 35.3 on \LSATA, with the largest gains on \textsc{Neither/Nor}. Direct prompting stays below 25 Macro-F1 on \textsc{Neither/Nor} for all models except \texttt{Llama-3.3-70B} with CoT on \LSATA~(28.9), so the failure is not specific to small models. For \texttt{Ministral-8B} and \texttt{Llama-3.3-70B}, independent true--false scoring is competitive with paired MC.

\paragraph{Performance Across Logical Operators}
The gains from structured inference are concentrated on \textsc{Or}, \textsc{Neither/Nor}, and \textsc{Mixed}, with the largest improvement on \textsc{Neither/Nor}. Adding few-shot demonstrations shifts answer-position preferences without improving \textsc{Neither/Nor} performance, suggesting that position bias alone does not explain these failures (App.~\ref{app:position-bias}). %The strongest direct-prompting baseline reaches only 13.9 macro-F1 on \LCQA~and 15.2 on \LSATA, whereas paired multiple-choice structured inference raises these to 75.1 and 71.9, respectively, and relative to 76.8 and 73.4. 
The recovery on \textsc{Neither/Nor} suggests that models retain useful evidence about the individual atomic answers even when they fail to combine two negative judgments correctly during direct compound-answer prediction. On \textsc{And}, results are mixed: a slight drop on \LCQA~and a gain on \LSATA. These results indicate that explicit logical composition is most beneficial when the final decision requires handling alternatives, jointly rejecting atomic answers, or applying different operators across candidate options.

\paragraph{Effect of Global Inference} We add two decomposition baselines that compose the same atomic scores locally, either by multiplying them as independent probabilities or by thresholding them and applying each operator's truth table. The ILP improves over them by 3.6 and 2.5 points on \LCQA-HV and \LSATA, since local composition often yields no unique valid option (App.~\ref{app:local-baselines}).

\paragraph{Effects of Calibration}
Calibration improves both the reliability of the atomic evidence scores and downstream compound prediction. As shown in Table~\ref{tab:atomic-calibration}, Platt scaling, isotonic calibration, and relative calibration all reduce atomic Brier score and log loss on both benchmarks, with relative calibration performing best on both metrics. %This indicates that incorporating within-instance information improves the calibration of local evidence scores. 
The reduction is proportionally larger for log loss than for Brier score, particularly on \LSATA, indicating that raw scores are overconfident rather than merely out of order. 

The downstream gains are smaller than the atomic improvements, since calibration changes the final prediction only when it alters the relative evidence among competing feasible assignments. Calibration helps most in \textsc{Mixed}, where relative calibration raises Macro-F1 by 4.9 on \LCQA-HV and 11.2 on \LSATA, against 3.4 and 9.5 for Platt scaling. When all four options share an operator, a systematic bias in the atomic scores shifts them equally and it largely cancels. In \textsc{Mixed}, options impose opposing demands, since \textsc{And} and \textsc{Or} require atoms to be accepted while \textsc{Neither/Nor} requires them to be rejected, so the same bias favors one operator over another. %Platt scaling and isotonic calibration apply a single global mapping and can only partly correct this, whereas relative calibration's features are defined against the other atoms in the instance.
Relative calibration corrects this better, since its features are defined against the other atoms in the instance.

\paragraph{Error Analysis} When the inference layer is provided with gold atomic statuses, accuracy reaches $1.00$ on both benchmarks. This follows by construction, as each instance has exactly one valid option, and the ILP encodes the operator semantics exactly. The informative quantity is thus how many atomic errors survive composition. Atomic accuracy is 0.830 on \LCQA-HV and 0.824 on \LSATA, against a compound accuracy of 0.758 and 0.723.

%An oracle analysis shows that the remaining errors originate primarily in the atomic layer rather than in the logical formulation. 
Qualitative analysis reveals different sources of atomic error across the benchmarks. On \LCQA, errors often involve broad interpretations of open-ended commonsense questions or insufficient attention to modifiers such as \emph{uncommon}. On \LSATA, the model often selects the label matching a passage's main topic, rejecting other labels that also apply. The logical operators then determine how these errors propagate: \textsc{Or} can tolerate an incorrect atomic judgment when another atomic answer remains supported, whereas \textsc{And} and \textsc{Neither/Nor} can be invalidated by a single incorrect assignment. \textsc{Mixed} instances are especially sensitive because the same atomic assignment can affect options governed by different operators. A full analysis and examples are provided in Appendix~\ref{app:error-analysis}.

\begin{table}[t]
\centering
\scriptsize
\setlength{\tabcolsep}{4pt}
\begin{tabular}{lrrrr}
\toprule
& \multicolumn{2}{c}{\textbf{LCQA-HV}} & \multicolumn{2}{c}{\textbf{LSATA}} \\
\cmidrule(lr){2-3} \cmidrule(lr){4-5}
\textbf{Method} & \textbf{Brier} $\downarrow$ & \textbf{LL} $\downarrow$ & \textbf{Brier} $\downarrow$ & \textbf{LL} $\downarrow$ \\
\midrule
Uncalibrated & 0.192 & 0.637 & 0.190 & 0.814 \\
Platt        & 0.168 & 0.508 & 0.149 & 0.457 \\
Isotonic     & 0.168 & 0.515 & 0.148 & 0.453 \\
Relative     & \textbf{0.146} & \textbf{0.457} & \textbf{0.145} & \textbf{0.444} \\
\bottomrule
\end{tabular}
\caption{\small{Atomic calibration results (Brier score and log loss, LL). Lower is better for both.}}
\label{tab:atomic-calibration}
\end{table}

% \begin{table}[t]
% \centering
% \scriptsize
% \begin{tabular}{llrr}
% \toprule
% \textbf{Benchmark} &
% \textbf{Method} &
% \textbf{Brier} $\downarrow$ &
% \textbf{Log loss} $\downarrow$ \\
% \midrule
% \multirow{4}{*}{LSATA}
% & Uncalibrated & 0.1896 & 0.8141 \\
% & Platt & 0.1493 & 0.4567 \\
% & Isotonic & 0.1483 & 0.4531 \\
% & Relative & \textbf{0.1449} & \textbf{0.4438} \\
% \midrule
% \multirow{4}{*}{LCQA-HV}
% & Uncalibrated & 0.1919 & 0.6368 \\
% & Platt & 0.1681 & 0.5076 \\
% & Isotonic & 0.1680 & 0.5145 \\
% & Relative & \textbf{0.1464} & \textbf{0.4567} \\
% \bottomrule
% \end{tabular}
% \caption{\small{Atomic calibration results. Brier score is the mean squared error between the predicted probability and the gold binary status; log loss is the negative log-likelihood of that status. Lower is better for both.}}
% \label{tab:atomic-calibration}
% \end{table}
\section{Conclusions and Future Work}

We study reasoning over compound options, where two atoms are joined by \textsc{And}, \textsc{Or}, or \textsc{Neither/Nor}. Direct prediction requires evaluating atomic answers and composing them in a single step. We propose a framework that separates these stages by eliciting local evidence for opposing hypotheses about each atom and combining it through globally constrained inference. We introduce relative calibration, which incorporates how each atomic score compares with other scores in the same instance. Across \LCQA~and \LSATA, structured inference substantially outperforms direct prompting, with largest improvements on \textsc{Neither/Nor}, and relative calibration gives targeted gains, mainly on \textsc{Mixed} and \textsc{Neither/Nor}. %These findings show that separating local atomic evaluation from explicit logical composition can reduce errors that arise during direct compound-answer prediction. 
More broadly, these results suggest that some failures on logical reasoning tasks reflect difficulties in composing local judgments, rather than only the absence of knowledge. 

Future work could extend the framework to options with more than two atomic answers, structures such as implication, exclusive disjunction, and nested expressions, settings where atomic answers and operators must be extracted from less structured text, and other reasoning benchmarks. A further direction is to replace inference with a probabilistic global formulation, so that atomic uncertainty propagates to compound prediction instead of being discarded at assignment time, yielding a distribution over options rather than a single choice.
\section*{Limitations}
The formulation and the two benchmarks assume exactly two atomic answers per option, exactly one valid option, and three operators. The results therefore do not establish that the same gains will hold for longer logical expressions, operators such as implication and exclusive disjunction, or tasks that permit multiple valid answers or none. The ILP is a flexible framework: additional operators, more atoms, or different answer-validity assumptions can be incorporated through different constraints, which we leave to future work. The framework depends on the quality of the atomic evidence supplied to the inference layer. Errors in commonsense interpretation, passage grounding, or source annotations can therefore propagate to the final prediction even when the logical constraints are applied correctly. \LCQA~may additionally contain questions with several plausible commonsense interpretations, while \LSATA~inherits the label definitions and domain coverage of SATA-Bench. Our primary scorer requires log-probabilities of the answer tokens, which many proprietary interfaces do not expose; our generation-based alternative avoids this but is more expensive, as it makes more calls, and at the same time performs worse.

\section*{Ethical Considerations}
This work uses publicly available source datasets, with human annotation limited to a small quality audit of LOGICAL-SATA. Nevertheless, the benchmarks and model outputs may reflect biases, ambiguities, or annotation errors present in their source datasets. The proposed framework improves consistency under explicit logical constraints, but it does not by itself guarantee that the underlying atomic judgments are correct. It should therefore not be interpreted as providing reliable logical guarantees for high-stakes applications such as medical, legal, or financial decision-making.

We used generative AI assistance in accordance with the ACL Policy on Publication Ethics. Its use was limited to language editing and compression, manuscript organization, \LaTeX{} formatting, and figure preparation. All output was reviewed, verified and further edited by the authors.
%\section*{Acknowledgements}

\bibliography{custom}

\appendix
\section*{Appendix}
\label{app: appendix}

\section{Implementation Details}
\label{app:implementation-details}
We use \texttt{Llama-3.1-8B-Instruct}, \texttt{Llama-3.2-1B-Instruct}, \texttt{Llama-3.2-3B-Instruct}, and \texttt{Llama-3.3-70B-Instruct} (4-bit AWQ) \citep{grattafiori2024llama}, and \texttt{Ministral-8B-Instruct-2410}. Direct prompting (0- to 3-shot) extracts the first-token log-probabilities of the option letters A--D. Predictions are sampled from the first-token distribution at temperature $0.7$ over five seeds. Chain-of-thought answers are parsed from the generated final line, since the answer follows the reasoning chain. All first-token direct-prompting predictions obtained were valid with no fallback requests. Atomic confidence scores are generated at a temperature of $0.7$. Structured-inference results are averaged over five runs. Generation sampling uses five generations per atomic answer.

For \LSATA, source questions are partitioned before constructing the four operator variants, ensuring that all variants of a source question remain in the same split. Calibration uses the full \LSATA~training set of $2{,}400$ instances. For \LCQA, we sample $2{,}400$ training instances, balanced across the four logical settings, to match the \LSATA~calibration-set size. Global inference is performed using Gurobi Optimizer 13.0.2. Experiments are run on an NVIDIA A100 GPU.

\paragraph{Calibration Implementation}
We fit all calibrators using atomic confidence scores and ground-truth binary labels from the respective benchmark training splits. Platt scaling uses logistic regression over the logit-transformed positive scores, while isotonic calibration fits a non-decreasing mapping with outputs bounded to $[0,1]$ and out-of-range inputs clipped to the fitted boundaries.

For relative calibration, we use the balanced training samples described above. We standardize the four features defined in Sec.~\ref{subsec:score_calibration} and fit an $\ell_2$-regularized logistic regression. The regularization parameter is selected by minimizing log loss on a held-out validation set, yielding $C=0.03$. We then fit the calibrator on the full calibration training set using L-BFGS with a maximum of 2,000 iterations.

\paragraph{Inference Cost}
\label{app:inference-cost}

Direct prompting requires an average of 1 model request per question. Under paired multiple-choice confidence scoring, each unique atomic answer requires one model request. Across both benchmarks, this corresponds to an average of 5.61 model requests per question. Calibration and ILP do not require additional model requests.

\section{Additional Results}
\label{app:additional-lcqa-results}

Table~\ref{tab:lcsqa-nv} reports Macro-F1 on the non-validated \LCQA~split.

\begin{table}[t]
\centering
\scriptsize
\setlength{\tabcolsep}{2.5pt}
\renewcommand{\arraystretch}{1.05}
\resizebox{\columnwidth}{!}{%
\begin{tabular}{lccccc}
\toprule
\textbf{Method}
& \textbf{\textsc{And}}
& \textbf{\textsc{Or}}
& \textbf{\textsc{NN}}
& \textbf{\textsc{Mix}}
& \textbf{Overall} \\
\midrule
\multicolumn{6}{l}{\textit{Direct prompting}} \\
0-shot
& $66.0^{\pm2.5}$
& $57.1^{\pm2.2}$
& $13.8^{\pm2.2}$
& $44.9^{\pm2.4}$
& $45.7^{\pm1.0}$ \\
1-shot
& $71.7^{\pm1.5}$
& $59.3^{\pm1.2}$
& $7.9^{\pm1.5}$
& $44.8^{\pm1.1}$
& $46.1^{\pm0.2}$ \\
2-shot
& $73.1^{\pm1.0}$
& $59.8^{\pm1.3}$
& $7.6^{\pm1.2}$
& $46.8^{\pm1.5}$
& $47.2^{\pm0.3}$ \\
3-shot
& $73.5^{\pm1.5}$
& $59.3^{\pm0.8}$
& $7.8^{\pm2.0}$
& $47.3^{\pm1.3}$
& $47.3^{\pm0.4}$ \\
CoT (0-shot)
& $66.8^{\pm2.8}$
& $65.3^{\pm2.1}$
& $11.7^{\pm1.9}$
& $48.6^{\pm1.2}$
& $48.3^{\pm0.7}$ \\
\midrule
\multicolumn{6}{l}{\textit{Structured inference}} \\
Paired MC
& $73.7^{\pm0.2}$
& $85.1^{\pm0.0}$
& $77.9^{\pm0.2}$
& $67.9^{\pm0.0}$
& $76.2^{\pm0.1}$ \\
\quad + Platt
& $74.8^{\pm0.4}$
& $85.1^{\pm0.0}$
& $77.6^{\pm0.0}$
& $68.2^{\pm0.0}$
& $76.5^{\pm0.1}$ \\
\quad + Isotonic
& $74.9^{\pm0.2}$
& $\mathbf{86.0^{\pm0.0}}$
& $76.4^{\pm0.0}$
& $68.0^{\pm0.2}$
& $76.4^{\pm0.0}$ \\
\quad + Relative
& $74.0^{\pm0.2}$
& $85.1^{\pm0.0}$
& $\mathbf{78.0^{\pm0.0}}$
& $\mathbf{74.3^{\pm0.0}}$
& $\mathbf{77.9^{\pm0.1}}$ \\
\bottomrule
\end{tabular}%
}
\caption{Macro-F1 on the non-validated (NV) split of \LCQA. NN = \textsc{Neither/Nor}.}
\label{tab:lcsqa-nv}
\end{table}

\subsection{Full Accuracy Results}
\label{app:accuracy-results}

Tables~\ref{tab:lcsqa-accuracy} and~\ref{tab:lsata-accuracy} report accuracy results for the direct-prompting and structured-inference configurations. 

\begin{table*}[t]
\centering
\small
\setlength{\tabcolsep}{4pt}
\renewcommand{\arraystretch}{1.1}
\resizebox{\textwidth}{!}{%
\begin{tabular}{lcccccccccc}
\toprule
& \multicolumn{2}{c}{\textsc{And}}
& \multicolumn{2}{c}{\textsc{Or}}
& \multicolumn{2}{c}{\textsc{NN}}
& \multicolumn{2}{c}{\textsc{Mixed}}
& \multicolumn{2}{c}{\textbf{Overall}} \\
\cmidrule(lr){2-3}
\cmidrule(lr){4-5}
\cmidrule(lr){6-7}
\cmidrule(lr){8-9}
\cmidrule(lr){10-11}
\textbf{Method}
& \textbf{NV} & \textbf{HV}
& \textbf{NV} & \textbf{HV}
& \textbf{NV} & \textbf{HV}
& \textbf{NV} & \textbf{HV}
& \textbf{NV} & \textbf{HV} \\
\midrule
\multicolumn{11}{l}{\textit{Direct prompting}} \\
0-shot
& $65.9^{\pm2.4}$ & $68.5^{\pm0.7}$
& $57.1^{\pm2.2}$ & $54.6^{\pm2.4}$
& $14.8^{\pm2.1}$ & $13.0^{\pm1.1}$
& $44.9^{\pm2.4}$ & $46.6^{\pm1.6}$
& $45.7^{\pm1.0}$ & $45.7^{\pm1.0}$ \\
1-shot
& $71.8^{\pm1.6}$ & $72.1^{\pm1.0}$
& $59.6^{\pm1.1}$ & $56.7^{\pm1.7}$
& $8.4^{\pm1.6}$ & $6.9^{\pm1.2}$
& $44.8^{\pm1.3}$ & $45.0^{\pm1.8}$
& $46.1^{\pm0.2}$ & $45.2^{\pm1.1}$ \\
2-shot
& $73.1^{\pm0.9}$ & $74.7^{\pm1.6}$
& $60.3^{\pm1.1}$ & $60.3^{\pm2.1}$
& $9.0^{\pm1.1}$ & $8.6^{\pm2.3}$
& $46.8^{\pm1.5}$ & $44.6^{\pm2.4}$
& $47.3^{\pm0.2}$ & $47.1^{\pm0.9}$ \\
3-shot
& $73.7^{\pm1.5}$ & $74.2^{\pm1.3}$
& $59.5^{\pm0.9}$ & $61.8^{\pm1.2}$
& $9.0^{\pm1.9}$ & $8.4^{\pm1.6}$
& $47.4^{\pm1.3}$ & $45.0^{\pm2.2}$
& $47.4^{\pm0.4}$ & $47.3^{\pm0.6}$ \\
CoT (0-shot)
& $66.5^{\pm2.8}$ & $69.9^{\pm1.8}$
& $65.0^{\pm2.2}$ & $62.3^{\pm0.7}$
& $12.1^{\pm2.0}$ & $14.3^{\pm0.8}$
& $48.4^{\pm1.2}$ & $45.8^{\pm2.5}$
& $48.0^{\pm0.8}$ & $48.1^{\pm0.5}$ \\
\midrule
\multicolumn{11}{l}{\textit{Structured inference}} \\
Paired MC
& $73.9^{\pm0.2}$ & $72.4^{\pm0.0}$
& $85.2^{\pm0.0}$ & $85.2^{\pm0.0}$
& $77.8^{\pm0.2}$ & $75.2^{\pm0.0}$
& $68.0^{\pm0.0}$ & $70.4^{\pm0.0}$
& $76.2^{\pm0.1}$ & $75.8^{\pm0.0}$ \\
+ Platt calibration
& $75.0^{\pm0.4}$ & $72.0^{\pm0.0}$
& $85.2^{\pm0.0}$ & $84.4^{\pm0.0}$
& $77.6^{\pm0.0}$ & $75.6^{\pm0.0}$
& $68.4^{\pm0.0}$ & $73.8^{\pm0.2}$
& $76.6^{\pm0.1}$ & $76.4^{\pm0.1}$ \\
+ Isotonic calibration
& $75.1^{\pm0.2}$ & $72.5^{\pm0.3}$
& $86.1^{\pm0.0}$ & $83.5^{\pm0.2}$
& $76.4^{\pm0.0}$ & $75.2^{\pm0.0}$
& $68.2^{\pm0.2}$ & $71.7^{\pm0.3}$
& $76.5^{\pm0.0}$ & $75.7^{\pm0.2}$ \\
+ Relative calibration
& $74.2^{\pm0.2}$ & $71.8^{\pm0.2}$
& $85.2^{\pm0.0}$ & $84.0^{\pm0.0}$
& $78.0^{\pm0.0}$ & $76.8^{\pm0.0}$
& $\mathbf{74.4^{\pm0.0}}$ & $\mathbf{75.2^{\pm0.0}}$
& $\mathbf{78.0^{\pm0.1}}$ & $\mathbf{77.0^{\pm0.1}}$ \\
\bottomrule
\end{tabular}
}
\caption{
Accuracy on the non-validated (NV) and human-validated
(HV) subsets of \LCQA{}.
}
\label{tab:lcsqa-accuracy}
\end{table*}

\begin{table}[t]
\centering
\small
\setlength{\tabcolsep}{5pt}
\renewcommand{\arraystretch}{1.1}
\resizebox{\columnwidth}{!}{%
\begin{tabular}{lccccc}
\toprule
\textbf{Method}
& \textsc{And}
& \textsc{Or}
& \textsc{NN}
& \textsc{Mixed}
& \textbf{Overall} \\
\midrule
\multicolumn{6}{l}{\textit{Direct prompting}} \\
0-shot
& $69.0^{\pm1.1}$
& $60.8^{\pm1.6}$
& $16.0^{\pm2.1}$
& $38.6^{\pm1.1}$
& $46.1^{\pm0.9}$ \\
1-shot
& $71.0^{\pm1.0}$
& $60.7^{\pm0.8}$
& $11.2^{\pm1.0}$
& $35.8^{\pm1.1}$
& $44.7^{\pm0.3}$ \\
2-shot
& $69.6^{\pm0.8}$
& $60.1^{\pm1.5}$
& $10.0^{\pm1.7}$
& $38.1^{\pm1.2}$
& $44.5^{\pm0.7}$ \\
3-shot
& $68.4^{\pm0.9}$
& $59.6^{\pm1.0}$
& $11.0^{\pm1.9}$
& $36.6^{\pm1.5}$
& $43.9^{\pm1.1}$ \\
CoT (0-shot)
& $70.5^{\pm1.6}$
& $65.7^{\pm1.3}$
& $15.1^{\pm2.2}$
& $39.3^{\pm2.7}$
& $47.7^{\pm1.4}$ \\
\midrule
\multicolumn{6}{l}{\textit{Structured inference}} \\
Paired MC
& $73.6^{\pm0.0}$
& $82.4^{\pm0.0}$
& $72.0^{\pm0.0}$
& $61.0^{\pm0.0}$
& $72.3^{\pm0.0}$ \\
+ Platt calibration
& $74.8^{\pm0.0}$
& $82.6^{\pm0.0}$
& $73.0^{\pm0.0}$
& $70.4^{\pm0.0}$
& $75.2^{\pm0.0}$ \\
+ Isotonic calibration
& $74.6^{\pm0.0}$
& $\mathbf{83.8^{\pm0.0}}$
& $73.0^{\pm0.0}$
& $69.8^{\pm0.0}$
& $75.3^{\pm0.0}$ \\
+ Relative calibration
& $74.4^{\pm0.0}$
& $82.4^{\pm0.0}$
& $\mathbf{73.4^{\pm0.0}}$
& $\mathbf{72.2^{\pm0.0}}$
& $\mathbf{75.6^{\pm0.0}}$ \\
\bottomrule
\end{tabular}%
}
\caption{
Accuracy on the \LSATA{} test set.
}
\label{tab:lsata-accuracy}
\end{table}

\subsection{Additional Models}
\label{app:other-models}
We repeat the experiments with four additional models, Llama-3.2-1B-Instruct, Llama-3.2-3B-Instruct, Llama-3.3-70B-Instruct (4-bit AWQ), and Ministral-8B-Instruct-2410 on \LCQA-HV and \LSATA~(Tables~\ref{tab:lcqa-hv-full} and~\ref{tab:lsata-full}). Structured inference with paired MC improves overall Macro-F1 over the best direct-prompting baseline for every model. The gain is largest on \textsc{Neither/Nor} (31.7 to 61.2 and 32.4 to 60.4 points). The gains are consistent across models on \textsc{Or}, \textsc{Neither/Nor}, and \textsc{Mixed}, whereas on \textsc{And} the effect varies by model.

Paired MC gives the best overall results for the 1B and 3B models. For the 70B model, independent true--false scoring is comparable on \LCQA-HV (78.2 vs.\ 77.8) and a little higher on \LSATA~(84.0 vs.\ 81.4), and for Ministral-8B the two methods are on par.

\begin{table*}[p]
\centering
\scriptsize
\setlength{\tabcolsep}{3pt}
\renewcommand{\arraystretch}{1.0}
\begin{minipage}{0.49\textwidth}
\centering
\resizebox{\textwidth}{!}{%
\begin{tabular}{llccccc}
\toprule
\textbf{Model} & \textbf{Method} & \textbf{AND} & \textbf{OR} & \textbf{NN} & \textbf{MIX} & \textbf{All} \\
\midrule
\multirow{9}{*}{Llama-3.2-1B}
& 0-shot          & $49.8^{\pm1.7}$ & $36.4^{\pm1.0}$ & $11.4^{\pm0.8}$ & $34.7^{\pm1.9}$ & $33.4^{\pm0.8}$ \\
& 1-shot          & $48.9^{\pm3.4}$ & $39.3^{\pm2.7}$ & $15.8^{\pm2.5}$ & $33.3^{\pm3.8}$ & $34.4^{\pm0.7}$ \\
& 2-shot          & $\mathbf{51.3}^{\pm2.9}$ & $39.8^{\pm2.7}$ & $14.8^{\pm2.0}$ & $31.4^{\pm1.1}$ & $34.7^{\pm1.3}$ \\
& 3-shot          & $49.2^{\pm2.4}$ & $40.5^{\pm3.4}$ & $14.5^{\pm2.9}$ & $32.6^{\pm2.1}$ & $34.4^{\pm1.6}$ \\
& CoT             & $43.0^{\pm3.0}$ & $36.1^{\pm2.2}$ & $16.0^{\pm2.2}$ & $32.3^{\pm1.9}$ & $32.2^{\pm0.9}$ \\
& Paired MC       & $50.7^{\pm0.2}$ & $\mathbf{68.7}^{\pm0.0}$ & $\mathbf{47.7}^{\pm0.2}$ & $\mathbf{42.7}^{\pm0.3}$ & $\mathbf{52.5}^{\pm0.1}$ \\
& Ind.\ T/F       & $33.1^{\pm0.0}$ & $58.5^{\pm0.2}$ & $35.6^{\pm0.5}$ & $37.2^{\pm0.0}$ & $41.2^{\pm0.1}$ \\
& Gen.\ samp.     & $37.3^{\pm1.9}$ & $61.3^{\pm2.3}$ & $38.8^{\pm1.2}$ & $38.9^{\pm2.0}$ & $44.1^{\pm1.2}$ \\
& Verbalized      & $27.8^{\pm2.6}$ & $52.9^{\pm2.2}$ & $27.4^{\pm3.4}$ & $35.8^{\pm1.0}$ & $36.1^{\pm1.6}$ \\
\midrule
\multirow{9}{*}{Llama-3.2-3B}
& 0-shot          & $63.7^{\pm1.7}$ & $46.1^{\pm1.9}$ & $13.4^{\pm1.4}$ & $40.9^{\pm1.5}$ & $41.1^{\pm1.0}$ \\
& 1-shot          & $70.2^{\pm1.5}$ & $53.6^{\pm0.8}$ & $9.2^{\pm2.1}$  & $45.2^{\pm2.5}$ & $44.7^{\pm1.0}$ \\
& 2-shot          & $69.2^{\pm1.9}$ & $51.9^{\pm1.8}$ & $8.7^{\pm1.6}$  & $45.0^{\pm1.7}$ & $44.0^{\pm1.1}$ \\
& 3-shot          & $69.3^{\pm0.8}$ & $53.8^{\pm1.1}$ & $8.6^{\pm1.5}$  & $44.2^{\pm1.9}$ & $44.3^{\pm0.9}$ \\
& CoT             & $69.6^{\pm0.9}$ & $60.1^{\pm1.2}$ & $11.4^{\pm2.5}$ & $46.9^{\pm1.7}$ & $47.0^{\pm1.1}$ \\
& Paired MC       & $\mathbf{71.8}^{\pm0.4}$ & $\mathbf{85.1}^{\pm0.4}$ & $\mathbf{71.9}^{\pm0.0}$ & $\mathbf{68.8}^{\pm0.0}$ & $\mathbf{74.4}^{\pm0.1}$ \\
& Ind.\ T/F       & $57.6^{\pm0.2}$ & $75.5^{\pm0.2}$ & $60.7^{\pm0.2}$ & $45.4^{\pm0.2}$ & $59.8^{\pm0.1}$ \\
& Gen.\ samp.     & $64.6^{\pm1.4}$ & $79.8^{\pm2.0}$ & $67.4^{\pm2.2}$ & $64.3^{\pm1.9}$ & $69.0^{\pm0.9}$ \\
& Verbalized      & $35.1^{\pm4.4}$ & $55.3^{\pm2.4}$ & $39.5^{\pm3.5}$ & $42.2^{\pm2.4}$ & $43.1^{\pm1.5}$ \\
\midrule
\multirow{9}{*}{Llama-3.1-8B}
& 0-shot          & $68.6^{\pm0.7}$ & $54.4^{\pm2.4}$ & $12.1^{\pm1.3}$ & $46.3^{\pm1.6}$ & $45.7^{\pm1.0}$ \\
& 1-shot          & $72.1^{\pm1.0}$ & $56.6^{\pm1.7}$ & $6.6^{\pm1.1}$  & $44.7^{\pm1.7}$ & $45.2^{\pm1.1}$ \\
& 2-shot          & $\mathbf{74.9}^{\pm1.7}$ & $60.2^{\pm2.3}$ & $7.7^{\pm2.1}$  & $43.9^{\pm2.4}$ & $47.0^{\pm0.9}$ \\
& 3-shot          & $74.3^{\pm1.3}$ & $61.8^{\pm1.3}$ & $7.5^{\pm1.6}$  & $44.5^{\pm2.1}$ & $47.3^{\pm0.6}$ \\
& CoT             & $70.1^{\pm1.7}$ & $62.3^{\pm0.7}$ & $13.9^{\pm0.7}$ & $46.1^{\pm2.7}$ & $48.3^{\pm0.5}$ \\
& Paired MC       & $72.4^{\pm0.0}$ & $\mathbf{85.2}^{\pm0.0}$ & $\mathbf{75.1}^{\pm0.0}$ & $\mathbf{70.3}^{\pm0.0}$ & $\mathbf{75.8}^{\pm0.0}$ \\
& Ind.\ T/F       & $70.8^{\pm0.0}$ & $80.8^{\pm0.0}$ & $73.6^{\pm0.0}$ & $\mathbf{70.3}^{\pm0.0}$ & $73.9^{\pm0.0}$ \\
& Gen.\ samp.     & $69.5^{\pm0.8}$ & $81.1^{\pm1.1}$ & $69.1^{\pm2.7}$ & $67.2^{\pm0.6}$ & $71.7^{\pm0.7}$ \\
& Verbalized      & $49.7^{\pm2.6}$ & $66.1^{\pm1.4}$ & $46.4^{\pm2.8}$ & $51.5^{\pm3.5}$ & $53.5^{\pm0.7}$ \\
\midrule
\multirow{9}{*}{Llama-3.3-70B}
& 0-shot          & $81.6^{\pm1.7}$ & $72.0^{\pm1.3}$ & $17.1^{\pm1.6}$ & $55.3^{\pm1.0}$ & $56.6^{\pm0.5}$ \\
& 1-shot          & $81.8^{\pm0.9}$ & $78.0^{\pm1.6}$ & $12.1^{\pm1.6}$ & $54.6^{\pm1.5}$ & $57.0^{\pm0.8}$ \\
& 2-shot          & $82.3^{\pm1.3}$ & $77.3^{\pm1.2}$ & $9.7^{\pm1.1}$  & $52.6^{\pm1.0}$ & $55.7^{\pm1.0}$ \\
& 3-shot          & $82.7^{\pm0.9}$ & $78.8^{\pm0.9}$ & $9.6^{\pm1.4}$  & $52.6^{\pm0.2}$ & $56.0^{\pm0.6}$ \\
& CoT             & $\mathbf{84.3}^{\pm1.7}$ & $73.4^{\pm1.3}$ & $23.2^{\pm1.6}$ & $56.9^{\pm0.6}$ & $59.5^{\pm0.9}$ \\
& Paired MC       & $78.3^{\pm0.3}$ & $\mathbf{87.3}^{\pm0.2}$ & $74.9^{\pm0.2}$ & $70.5^{\pm0.3}$ & $77.8^{\pm0.2}$ \\
& Ind.\ T/F       & $77.2^{\pm0.0}$ & $86.1^{\pm0.0}$ & $\mathbf{75.4}^{\pm0.0}$ & $\mathbf{74.0}^{\pm0.0}$ & $\mathbf{78.2}^{\pm0.0}$ \\
& Gen.\ samp.     & $75.0^{\pm1.3}$ & $82.8^{\pm0.9}$ & $72.8^{\pm1.3}$ & $69.3^{\pm1.8}$ & $75.0^{\pm0.5}$ \\
& Verbalized      & $70.8^{\pm3.6}$ & $80.6^{\pm1.8}$ & $71.3^{\pm2.5}$ & $67.1^{\pm3.0}$ & $72.4^{\pm1.5}$ \\
\midrule
\multirow{9}{*}{Ministral-8B}
& 0-shot          & $59.7^{\pm4.1}$ & $50.2^{\pm1.6}$ & $16.3^{\pm0.9}$ & $43.7^{\pm2.7}$ & $42.6^{\pm0.4}$ \\
& 1-shot          & $64.1^{\pm3.1}$ & $55.4^{\pm2.3}$ & $10.7^{\pm1.8}$ & $44.2^{\pm1.2}$ & $43.7^{\pm0.6}$ \\
& 2-shot          & $67.0^{\pm3.0}$ & $59.1^{\pm2.3}$ & $11.5^{\pm2.1}$ & $44.2^{\pm1.0}$ & $45.6^{\pm1.1}$ \\
& 3-shot          & $66.8^{\pm1.9}$ & $58.9^{\pm2.5}$ & $11.3^{\pm1.8}$ & $44.1^{\pm1.6}$ & $45.4^{\pm0.5}$ \\
& CoT             & $63.1^{\pm3.2}$ & $59.0^{\pm2.5}$ & $14.1^{\pm2.7}$ & $44.2^{\pm2.7}$ & $45.2^{\pm1.3}$ \\
& Paired MC       & $73.8^{\pm0.4}$ & $86.4^{\pm0.2}$ & $\mathbf{75.5}^{\pm0.3}$ & $\mathbf{72.9}^{\pm0.2}$ & $77.1^{\pm0.0}$ \\
& Ind.\ T/F       & $\mathbf{75.6}^{\pm0.6}$ & $\mathbf{87.5}^{\pm0.5}$ & $74.2^{\pm0.3}$ & $72.3^{\pm0.2}$ & $\mathbf{77.4}^{\pm0.3}$ \\
& Gen.\ samp.     & $65.3^{\pm2.0}$ & $81.6^{\pm1.2}$ & $66.7^{\pm2.5}$ & $66.6^{\pm1.6}$ & $70.1^{\pm1.1}$ \\
& Verbalized      & $35.8^{\pm3.3}$ & $61.0^{\pm1.3}$ & $41.3^{\pm2.5}$ & $44.1^{\pm2.6}$ & $45.6^{\pm1.2}$ \\
\bottomrule
\end{tabular}%
}
\caption{Macro-F1 (mean $\pm$ std over five runs) on \LCQA-HV for additional models. NN = \textsc{Neither/Nor}.}
\label{tab:lcqa-hv-full}
\end{minipage}
\hfill
\begin{minipage}{0.49\textwidth}
\centering
\resizebox{\textwidth}{!}{%
\begin{tabular}{llccccc}
\toprule
\textbf{Model} & \textbf{Method} & \textbf{AND} & \textbf{OR} & \textbf{NN} & \textbf{MIX} & \textbf{All} \\
\midrule
\multirow{9}{*}{Llama-3.2-1B}
& 0-shot          & $\mathbf{51.9}^{\pm2.0}$ & $36.2^{\pm1.6}$ & $14.4^{\pm2.1}$ & $27.8^{\pm0.8}$ & $32.8^{\pm1.2}$ \\
& 1-shot          & $47.0^{\pm1.5}$ & $33.5^{\pm1.0}$ & $17.2^{\pm1.9}$ & $27.0^{\pm1.4}$ & $31.3^{\pm1.2}$ \\
& 2-shot          & $45.9^{\pm2.6}$ & $31.3^{\pm2.0}$ & $16.6^{\pm2.2}$ & $26.7^{\pm1.2}$ & $30.2^{\pm0.9}$ \\
& 3-shot          & $44.0^{\pm2.1}$ & $31.8^{\pm2.4}$ & $17.0^{\pm1.0}$ & $26.8^{\pm1.4}$ & $30.0^{\pm1.2}$ \\
& CoT             & $42.1^{\pm1.9}$ & $34.8^{\pm1.6}$ & $16.8^{\pm1.4}$ & $25.0^{\pm1.8}$ & $30.0^{\pm0.9}$ \\
& Paired MC       & $49.5^{\pm0.7}$ & $\mathbf{62.6}^{\pm0.9}$ & $\mathbf{49.6}^{\pm0.4}$ & $46.7^{\pm0.1}$ & $\mathbf{52.2}^{\pm0.3}$ \\
& Ind.\ T/F       & $27.6^{\pm0.3}$ & $49.9^{\pm0.4}$ & $25.3^{\pm0.3}$ & $47.4^{\pm0.2}$ & $37.6^{\pm0.1}$ \\
& Gen.\ samp.     & $38.9^{\pm2.2}$ & $58.9^{\pm1.8}$ & $45.3^{\pm2.0}$ & $\mathbf{47.9}^{\pm1.0}$ & $47.8^{\pm0.4}$ \\
& Verbalized      & $26.2^{\pm2.2}$ & $49.2^{\pm1.5}$ & $27.1^{\pm2.3}$ & $\mathbf{47.9}^{\pm1.5}$ & $37.6^{\pm0.8}$ \\
\midrule
\multirow{9}{*}{Llama-3.2-3B}
& 0-shot          & $63.4^{\pm1.5}$ & $49.5^{\pm2.0}$ & $12.2^{\pm1.0}$ & $33.3^{\pm2.1}$ & $39.9^{\pm1.4}$ \\
& 1-shot          & $67.1^{\pm1.0}$ & $53.0^{\pm2.4}$ & $8.1^{\pm1.4}$  & $34.4^{\pm1.1}$ & $40.8^{\pm0.9}$ \\
& 2-shot          & $68.2^{\pm0.4}$ & $53.9^{\pm2.7}$ & $7.6^{\pm1.4}$  & $34.4^{\pm1.1}$ & $41.2^{\pm1.0}$ \\
& 3-shot          & $67.6^{\pm1.0}$ & $52.6^{\pm1.4}$ & $8.2^{\pm0.8}$  & $33.7^{\pm0.9}$ & $40.7^{\pm0.4}$ \\
& CoT             & $68.4^{\pm0.5}$ & $59.3^{\pm1.7}$ & $9.4^{\pm0.9}$  & $36.5^{\pm1.6}$ & $43.5^{\pm0.9}$ \\
& Paired MC       & $\mathbf{74.7}^{\pm0.4}$ & $\mathbf{82.8}^{\pm0.6}$ & $\mathbf{70.2}^{\pm0.2}$ & $62.1^{\pm0.3}$ & $\mathbf{72.5}^{\pm0.2}$ \\
& Ind.\ T/F       & $68.4^{\pm0.4}$ & $78.5^{\pm0.6}$ & $65.4^{\pm0.6}$ & $58.7^{\pm0.4}$ & $67.8^{\pm0.2}$ \\
& Gen.\ samp.     & $68.5^{\pm1.2}$ & $77.5^{\pm1.1}$ & $65.7^{\pm1.3}$ & $\mathbf{62.7}^{\pm1.0}$ & $68.6^{\pm0.7}$ \\
& Verbalized      & $34.4^{\pm0.3}$ & $53.1^{\pm0.8}$ & $37.2^{\pm1.3}$ & $52.0^{\pm2.9}$ & $44.3^{\pm0.8}$ \\
\midrule
\multirow{9}{*}{Llama-3.1-8B}
& 0-shot          & $69.0^{\pm1.1}$ & $60.5^{\pm1.6}$ & $15.2^{\pm2.1}$ & $38.5^{\pm1.1}$ & $46.2^{\pm0.9}$ \\
& 1-shot          & $71.0^{\pm1.0}$ & $60.3^{\pm0.9}$ & $11.2^{\pm1.0}$ & $35.5^{\pm1.2}$ & $44.6^{\pm0.4}$ \\
& 2-shot          & $69.7^{\pm0.8}$ & $59.5^{\pm1.5}$ & $10.1^{\pm1.6}$ & $37.8^{\pm1.2}$ & $44.4^{\pm0.7}$ \\
& 3-shot          & $68.4^{\pm0.9}$ & $59.1^{\pm0.9}$ & $11.0^{\pm1.9}$ & $36.2^{\pm1.6}$ & $43.9^{\pm1.1}$ \\
& CoT             & $70.5^{\pm1.7}$ & $65.2^{\pm1.4}$ & $14.5^{\pm2.4}$ & $39.2^{\pm2.6}$ & $47.6^{\pm1.4}$ \\
& Paired MC       & $\mathbf{73.6}^{\pm0.0}$ & $\mathbf{82.2}^{\pm0.0}$ & $\mathbf{71.9}^{\pm0.0}$ & $60.9^{\pm0.0}$ & $\mathbf{72.2}^{\pm0.0}$ \\
& Ind.\ T/F       & $70.3^{\pm0.0}$ & $78.6^{\pm0.0}$ & $68.6^{\pm0.0}$ & $49.7^{\pm0.0}$ & $66.8^{\pm0.0}$ \\
& Gen.\ samp.     & $63.1^{\pm1.0}$ & $77.7^{\pm1.9}$ & $65.5^{\pm0.5}$ & $\mathbf{61.5}^{\pm0.5}$ & $66.9^{\pm0.6}$ \\
& Verbalized      & $41.6^{\pm1.2}$ & $55.7^{\pm1.0}$ & $41.6^{\pm2.5}$ & $53.1^{\pm0.9}$ & $48.0^{\pm0.8}$ \\
\midrule
\multirow{9}{*}{Llama-3.3-70B}
& 0-shot          & $85.5^{\pm0.5}$ & $75.6^{\pm0.5}$ & $18.9^{\pm0.9}$ & $53.7^{\pm0.9}$ & $58.7^{\pm0.4}$ \\
& 1-shot          & $85.2^{\pm0.6}$ & $77.1^{\pm0.6}$ & $8.3^{\pm1.0}$  & $52.5^{\pm1.2}$ & $56.0^{\pm0.4}$ \\
& 2-shot          & $84.3^{\pm0.6}$ & $76.7^{\pm0.7}$ & $6.7^{\pm0.8}$  & $50.9^{\pm1.2}$ & $54.7^{\pm0.4}$ \\
& 3-shot          & $84.1^{\pm0.5}$ & $76.7^{\pm1.1}$ & $9.8^{\pm1.9}$  & $48.9^{\pm1.6}$ & $55.1^{\pm0.4}$ \\
& CoT             & $\mathbf{86.3}^{\pm0.6}$ & $78.6^{\pm0.8}$ & $28.9^{\pm2.7}$ & $53.4^{\pm1.0}$ & $61.9^{\pm1.1}$ \\
& Paired MC       & $82.6^{\pm0.0}$ & $88.5^{\pm0.0}$ & $80.6^{\pm0.0}$ & $73.7^{\pm0.0}$ & $81.4^{\pm0.0}$ \\
& Ind.\ T/F       & $85.0^{\pm0.0}$ & $\mathbf{89.5}^{\pm0.0}$ & $\mathbf{82.8}^{\pm0.0}$ & $\mathbf{78.6}^{\pm0.0}$ & $\mathbf{84.0}^{\pm0.0}$ \\
& Gen.\ samp.     & $75.0^{\pm0.6}$ & $82.8^{\pm1.0}$ & $74.8^{\pm0.4}$ & $72.7^{\pm0.7}$ & $76.3^{\pm0.4}$ \\
& Verbalized      & $78.6^{\pm1.9}$ & $83.2^{\pm0.5}$ & $75.9^{\pm1.0}$ & $76.4^{\pm0.2}$ & $78.6^{\pm0.6}$ \\
\midrule
\multirow{9}{*}{Ministral-8B}
& 0-shot          & $64.6^{\pm1.2}$ & $51.4^{\pm2.3}$ & $17.4^{\pm2.4}$ & $35.7^{\pm0.4}$ & $42.4^{\pm1.3}$ \\
& 1-shot          & $68.5^{\pm1.0}$ & $59.4^{\pm2.4}$ & $12.3^{\pm1.8}$ & $36.5^{\pm1.4}$ & $44.3^{\pm0.9}$ \\
& 2-shot          & $69.2^{\pm1.3}$ & $58.7^{\pm2.8}$ & $12.6^{\pm2.7}$ & $37.1^{\pm2.2}$ & $44.5^{\pm1.2}$ \\
& 3-shot          & $68.1^{\pm1.4}$ & $57.8^{\pm3.0}$ & $12.4^{\pm1.7}$ & $35.7^{\pm1.5}$ & $43.6^{\pm1.1}$ \\
& CoT             & $67.1^{\pm1.4}$ & $53.5^{\pm2.3}$ & $17.3^{\pm1.6}$ & $37.1^{\pm1.0}$ & $44.0^{\pm0.8}$ \\
& Paired MC       & $\mathbf{81.6}^{\pm0.3}$ & $\mathbf{88.1}^{\pm0.2}$ & $\mathbf{77.8}^{\pm0.0}$ & $71.7^{\pm0.3}$ & $\mathbf{79.8}^{\pm0.1}$ \\
& Ind.\ T/F       & $80.7^{\pm0.3}$ & $86.1^{\pm0.0}$ & $76.2^{\pm0.0}$ & $\mathbf{74.4}^{\pm0.0}$ & $79.4^{\pm0.1}$ \\
& Gen.\ samp.     & $75.2^{\pm1.4}$ & $84.1^{\pm1.1}$ & $72.5^{\pm1.8}$ & $70.8^{\pm0.5}$ & $75.7^{\pm0.4}$ \\
& Verbalized      & $43.8^{\pm2.3}$ & $55.7^{\pm1.2}$ & $43.4^{\pm1.9}$ & $54.5^{\pm2.2}$ & $49.4^{\pm1.1}$ \\
\bottomrule
\end{tabular}%
}
\caption{Macro-F1 (mean $\pm$ std over five runs) on the \LSATA~test set for additional models. NN = \textsc{Neither/Nor}.}
\label{tab:lsata-full}
\end{minipage}
\end{table*}

\subsection{Generation-Based Direct Prompting}
\label{app:generation-direct}
% \new{
Tables~\ref{tab:gen-direct-lcqa} and~\ref{tab:gen-direct-lsata} report direct prompting, where the answer letter is sampled at temperature 0.7 and parsed. 
% }

\begin{table}[t]
\centering
\scriptsize
\setlength{\tabcolsep}{2.5pt}
\renewcommand{\arraystretch}{1.05}
\resizebox{\columnwidth}{!}{%
\begin{tabular}{lccccc}
\toprule
\textbf{Method}
& \textbf{\textsc{And}}
& \textbf{\textsc{Or}}
& \textbf{\textsc{NN}}
& \textbf{\textsc{Mix}}
& \textbf{All} \\
\midrule
0-shot
& $68.2^{\pm1.1}$
& $55.7^{\pm2.4}$
& $14.0^{\pm1.8}$
& $47.2^{\pm2.0}$
& $46.5^{\pm1.3}$ \\
1-shot
& $70.8^{\pm0.6}$
& $54.4^{\pm1.9}$
& $8.7^{\pm0.8}$
& $43.5^{\pm1.1}$
& $44.5^{\pm0.4}$ \\
2-shot
& $64.2^{\pm2.1}$
& $53.3^{\pm3.7}$
& $9.4^{\pm1.5}$
& $41.1^{\pm3.5}$
& $42.2^{\pm1.4}$ \\
3-shot
& $57.4^{\pm3.1}$
& $52.9^{\pm3.0}$
& $8.5^{\pm0.8}$
& $40.9^{\pm2.2}$
& $40.1^{\pm1.0}$ \\
\bottomrule
\end{tabular}%
}
\caption{Macro-F1 on the human-validated (HV) split of \LCQA~with generation-based direct prompting. NN = \textsc{Neither/Nor}.}
\label{tab:gen-direct-lcqa}
\end{table}

\begin{table}[t]
\centering
\scriptsize
\setlength{\tabcolsep}{2.5pt}
\renewcommand{\arraystretch}{1.05}
\resizebox{\columnwidth}{!}{%
\begin{tabular}{lccccc}
\toprule
\textbf{Method}
& \textbf{AND}
& \textbf{OR}
& \textbf{NN}
& \textbf{MIX}
& \textbf{All} \\
\midrule
0-shot
& $64.4^{\pm1.2}$
& $58.7^{\pm1.6}$
& $12.1^{\pm1.0}$
& $38.6^{\pm0.9}$
& $44.0^{\pm0.6}$ \\
1-shot
& $70.9^{\pm1.1}$
& $60.8^{\pm0.6}$
& $9.3^{\pm1.0}$
& $36.4^{\pm1.8}$
& $44.4^{\pm0.8}$ \\
2-shot
& $70.9^{\pm1.3}$
& $60.4^{\pm1.6}$
& $8.9^{\pm1.3}$
& $38.0^{\pm1.2}$
& $44.8^{\pm0.5}$ \\
3-shot
& $70.3^{\pm0.8}$
& $59.2^{\pm0.9}$
& $9.9^{\pm1.2}$
& $38.1^{\pm1.2}$
& $44.6^{\pm0.4}$ \\
\bottomrule
\end{tabular}%
}
\caption{Macro-F1 on the \LSATA~test set with generation-based direct prompting. NN = \textsc{Neither/Nor}.}
\label{tab:gen-direct-lsata}
\end{table}

\subsection{Direct-Prompting Position Bias}
\label{app:position-bias}

Table~\ref{tab:position-dist} shows the answer positions predicted by direct prompting on \textsc{Neither/Nor}. On \LCQA-HV, adding demonstrations shifts predictions from option A (39.9\% at 0-shot, 12.6\% at 3-shot) toward D (28.2\% to 44.2\%). On \LSATA, predictions become more evenly distributed with demonstrations, yet \textsc{Neither/Nor} performance remains low. These results therefore suggest that answer-position preferences do not resolve the poor performance. Models continue to struggle with \textsc{Neither/Nor} reasoning despite changes in their answer-position preferences. This persistent weakness indicates difficulties in composing negative judgments under the \textsc{Neither/Nor} operator.

\begin{table}[t]
\centering
\scriptsize
\setlength{\tabcolsep}{2.5pt}
\renewcommand{\arraystretch}{1.05}
\resizebox{\columnwidth}{!}{%
\begin{tabular}{lccccc}
\toprule
Dataset & Setting & A & B & C & D \\
\midrule
\multirow{4}{*}{\LCQA-HV}
 & 0-shot & 39.9 & 16.2 & 15.7 & 28.2 \\
 & 1-shot & 17.0 & 19.3 & 31.4 & 32.3 \\
 & 2-shot & 11.8 & 18.2 & 26.0 & 44.0 \\
 & 3-shot & 12.6 & 18.2 & 25.1 & 44.2 \\
\midrule
\multirow{4}{*}{\LSATA}
 & 0-shot & 36.2 & 21.3 & 14.5 & 28.0 \\
 & 1-shot & 21.8 & 22.9 & 31.2 & 24.2 \\
 & 2-shot & 23.5 & 22.0 & 29.3 & 25.2 \\
 & 3-shot & 22.6 & 20.5 & 28.3 & 28.6 \\
\bottomrule
\end{tabular}%
}
\caption{\small{Predicted answer-position distribution (\%) of direct prompting on \textsc{Neither/Nor}, averaged over five runs.}}
\label{tab:position-dist}
\end{table}

\section{Alternative Confidence Elicitation Strategies}
\label{app:elicitation}
In addition to paired multiple-choice confidence, we evaluate three alternative strategies that differ in how they obtain support for the positive and negative hypotheses. Each method produces non-negative support values
\[
r_C^{+}(a),\,
r_C^{-}(a),
\]
which we normalize to obtain raw evidence scores:
\begin{align*}
s_{C,\mathrm{raw}}^{+}(a)
&=
\frac{r_C^{+}(a)}
{r_C^{+}(a)+r_C^{-}(a)},
\\[4pt]
s_{C,\mathrm{raw}}^{-}(a)
&=
\frac{r_C^{-}(a)}
{r_C^{+}(a)+r_C^{-}(a)}.
\end{align*}

\subsection{Independent True--False Confidence}
We evaluate the positive and negative hypotheses independently using separate prompts. For each hypothesis, we extract the probability assigned to the \texttt{True} answer token:
\begin{align*}
r_C^{+}(a)
&=
\Pr\!\left(
\texttt{True}\mid C, h_C^{+}(a)
\right),
\\
r_C^{-}(a)
&=
\Pr\!\left(
\texttt{True}\mid C, h_C^{-}(a)
\right).
\end{align*}
The two independently obtained values are then normalized to form relative evidence scores.

\subsection{Generation Sampling}
We use the same paired prompt as in the main method but estimate confidence through repeated stochastic generation rather than token probabilities. We sample the model $N$ times and parse each generation $g_n$ as selecting either the positive hypothesis, represented by \texttt{A}, or the negative hypothesis, represented by \texttt{B}. The support values are their empirical selection frequencies:
\begin{align*}
r_C^{+}(a)
&=
\frac{1}{N}
\sum_{n=1}^{N}
\mathbb{I}[g_n=\texttt{A}],
\\
r_C^{-}(a)
&=
\frac{1}{N}
\sum_{n=1}^{N}
\mathbb{I}[g_n=\texttt{B}].
\end{align*}
Because every valid generation selects one of the two alternatives, these values already sum to one. The sampling parameters are reported in Appendix~\ref{app:implementation-details}.

\subsection{Verbalized Confidence}
We evaluate the positive and negative hypotheses in separate model calls and ask the model to report a numerical confidence score between $0$ and $10$. The parsed responses define $r_C^{+}(a)$ and $r_C^{-}(a)$, which are normalized to obtain the corresponding raw evidence scores. Unlike the other strategies, verbalized confidence relies on the model's self-reported numerical judgment rather than answer-token probabilities or repeated selections.

\subsection{Comparison of Confidence Elicitation Strategies}
\label{app:elicitation-results}

We compare four strategies for eliciting local evidence: paired multiple-choice confidence, independent true--false confidence, generation sampling, and verbalized confidence. Results are reported as mean $\pm$ standard deviation over five runs.

\begin{table*}[t]
\centering
\scriptsize
\setlength{\tabcolsep}{2.3pt}
\renewcommand{\arraystretch}{1.05}

\resizebox{\textwidth}{!}{%
\begin{tabular}{lcccccccccc}
\toprule
\textbf{Method}
& \multicolumn{2}{c}{\textbf{AND}}
& \multicolumn{2}{c}{\textbf{OR}}
& \multicolumn{2}{c}{\textbf{NN}}
& \multicolumn{2}{c}{\textbf{MIX}}
& \multicolumn{2}{c}{\textbf{Overall}} \\
\cmidrule(lr){2-3}
\cmidrule(lr){4-5}
\cmidrule(lr){6-7}
\cmidrule(lr){8-9}
\cmidrule(lr){10-11}
& NV F1 & HV F1
& NV F1 & HV F1
& NV F1 & HV F1
& NV F1 & HV F1
& NV F1 & HV F1 \\
\midrule

Paired MC
& $73.7 \pm 0.2$
& $\mathbf{72.4 \pm 0.0}$
& $\mathbf{85.1 \pm 0.0}$
& $\mathbf{85.2 \pm 0.0}$
& $\mathbf{77.9 \pm 0.2}$
& $\mathbf{75.1 \pm 0.0}$
& $\mathbf{67.9 \pm 0.0}$
& $\mathbf{70.3 \pm 0.0}$
& $\mathbf{76.2 \pm 0.1}$
& $\mathbf{75.8 \pm 0.0}$ \\

Independent T/F
& $\mathbf{74.7 \pm 0.0}$
& $70.8 \pm 0.0$
& $81.6 \pm 0.0$
& $80.8 \pm 0.0$
& $75.6 \pm 0.0$
& $73.6 \pm 0.0$
& $66.3 \pm 0.0$
& $\mathbf{70.3 \pm 0.0}$
& $74.6 \pm 0.0$
& $73.9 \pm 0.0$ \\

Generation sampling
& $70.7 \pm 1.7$
& $69.5 \pm 0.8$
& $84.3 \pm 0.6$
& $81.1 \pm 1.1$
& $71.5 \pm 2.0$
& $69.1 \pm 2.7$
& $66.1 \pm 0.8$
& $67.2 \pm 0.6$
& $73.2 \pm 1.0$
& $71.7 \pm 0.7$ \\

Verbalized confidence
& $52.5 \pm 2.6$
& $49.7 \pm 2.6$
& $68.5 \pm 1.5$
& $66.1 \pm 1.4$
& $54.7 \pm 4.3$
& $46.4 \pm 2.8$
& $51.3 \pm 2.8$
& $51.5 \pm 3.5$
& $56.8 \pm 1.5$
& $53.5 \pm 0.7$ \\

\bottomrule
\end{tabular}%
}

\caption{Macro-F1 for alternative confidence elicitation strategies on the non-validated (NV) and human-validated (HV) subsets of \LCQA. NN denotes \textsc{Neither/Nor}.}
\label{tab:elicitation-lcqa}
\end{table*}

\begin{table}[t]
\centering
\scriptsize
\setlength{\tabcolsep}{2.5pt}
\renewcommand{\arraystretch}{1.05}

\resizebox{\columnwidth}{!}{%
\begin{tabular}{lccccc}
\toprule
\textbf{Method}
& \textbf{AND}
& \textbf{OR}
& \textbf{NN}
& \textbf{MIX}
& \textbf{Overall} \\
\midrule

Paired MC
& $\mathbf{73.6 \pm 0.0}$
& $\mathbf{82.2 \pm 0.0}$
& $\mathbf{71.9 \pm 0.0}$
& $60.9 \pm 0.0$
& $\mathbf{72.2 \pm 0.0}$ \\

Independent T/F
& $70.3 \pm 0.0$
& $78.6 \pm 0.0$
& $68.6 \pm 0.0$
& $49.7 \pm 0.0$
& $66.8 \pm 0.0$ \\

Generation sampling
& $63.1 \pm 1.0$
& $77.7 \pm 1.9$
& $65.5 \pm 0.5$
& $\mathbf{61.5 \pm 0.5}$
& $66.9 \pm 0.6$ \\

Verbalized confidence
& $41.6 \pm 1.2$
& $55.7 \pm 1.0$
& $41.6 \pm 2.5$
& $53.1 \pm 0.9$
& $48.0 \pm 0.8$ \\

\bottomrule
\end{tabular}%
}

\caption{Macro-F1 for alternative confidence elicitation strategies on the \LSATA~test set.}
\label{tab:elicitation-lsata}
\end{table}

As shown in Tables~\ref{tab:elicitation-lcqa} and~\ref{tab:elicitation-lsata}, paired multiple-choice confidence achieves the strongest overall performance on both benchmarks. It improves overall Macro-F1 over independent true--false confidence from 74.6 to 76.2 on \LCQA-NV, from 73.9 to 75.8 on \LCQA-HV, and from 66.8 to 72.2 on \LSATA{}. This pattern suggests that directly contrasting the positive and negative hypotheses within the same prompt provides more useful local evidence than evaluating them independently.

Generation sampling performs below paired multiple-choice confidence overall and exhibits greater variability because its evidence scores are estimated from stochastic generations. It nevertheless slightly outperforms paired multiple choice in the \textsc{Mixed} setting on \LSATA. Verbalized confidence performs substantially worse across both benchmarks, indicating that self-reported numerical confidence provides less reliable evidence for globally constrained inference than answer-token probabilities or repeated model selections.

\subsection{Decomposition Baselines}
\label{app:local-baselines}

\begin{table}[t]
\centering
\resizebox{\columnwidth}{!}{%
\begin{tabular}{llccccc}
\toprule
Dataset & Method & \textsc{And} & \textsc{Or} & NN & Mix & All \\
\midrule
\multirow{3}{*}{\LCQA-HV}
 & Probabilistic composition  & $\mathbf{72.7}^{\pm0.2}$ & $74.2^{\pm0.2}$ & $74.5^{\pm0.2}$ & $67.4^{\pm0.2}$ & $72.2^{\pm0.1}$ \\
 & Boolean + score resolution & $\mathbf{72.7}^{\pm0.2}$ & $74.2^{\pm0.2}$ & $74.5^{\pm0.2}$ & $67.1^{\pm0.0}$ & $72.1^{\pm0.1}$ \\
 & Global ILP                 & $72.4^{\pm0.0}$ & $\mathbf{85.2}^{\pm0.0}$ & $\mathbf{75.1}^{\pm0.0}$ & $\mathbf{70.3}^{\pm0.0}$ & $\mathbf{75.8}^{\pm0.0}$ \\
\midrule
\multirow{3}{*}{\LSATA}
 & Probabilistic composition  & $\mathbf{74.5}^{\pm0.3}$ & $72.3^{\pm0.3}$ & $70.7^{\pm0.4}$ & $\mathbf{61.1}^{\pm0.4}$ & $69.7^{\pm0.3}$ \\
 & Boolean + score resolution & $\mathbf{74.5}^{\pm0.3}$ & $72.5^{\pm0.3}$ & $70.7^{\pm0.4}$ & $60.8^{\pm0.4}$ & $69.7^{\pm0.3}$ \\
 & Global ILP                 & $73.6^{\pm0.0}$ & $\mathbf{82.2}^{\pm0.0}$ & $\mathbf{71.9}^{\pm0.0}$ & $60.9^{\pm0.0}$ & $\mathbf{72.2}^{\pm0.0}$ \\
\bottomrule
\end{tabular}%
}
\caption{Macro-F1 (mean $\pm$ std over five runs) of decomposition baselines and global ILP inference. NN = \textsc{Neither/Nor}.}
\label{tab:local-baselines}
\end{table}
% \new{
To isolate the contribution of global inference, we compare our ILP framework with two decomposition baselines. Both use the same uncalibrated atomic scores as the ILP, but compose them locally, without a global objective or the exactly-one constraint.

\paragraph{Probabilistic composition} We compose the atomic scores directly into an option score and predict the highest-scoring option. For option $A_i$, with $s_1$ and $s_2$ denoting $s^+_{C,\mathrm{raw}}$ of its two atoms, the option score is
\begin{equation}
\small
\label{eq:prob-composition}
\psi_{\circ}(s_1, s_2) =
\begin{cases}
s_1 s_2, & \circ = \textsc{And} \\[2pt]
1 - (1 - s_1)(1 - s_2), & \circ = \textsc{Or} \\[2pt]
(1 - s_1)(1 - s_2), & \circ = \textsc{Neither/Nor}.
\end{cases}
\end{equation}
\paragraph{Boolean + score resolution} We threshold each atomic score at 0.5 and then apply corresponding Boolean truth independently to each option. We use a score-based rule as defined in Eq.~\ref{eq:prob-composition} when zero or multiple options remain valid.

\paragraph{Results} Table~\ref{tab:local-baselines} reports Macro-F1. The two baselines perform almost identical despite using different composition rules, and both improve on the best direct-prompting baselines (48.3 on \LCQA-HV and 47.6 on \LSATA) by about 24 and 22 points, respectively. Therefore decomposition accounts for most of the gain over direct prompting. The ILP adds additional 3.6 points on \LCQA-HV and 2.5 on \LSATA~over the stronger baseline. The gain comes mainly from \textsc{Or}, where Macro-F1 rises from 74.2 to 85.2 on \LCQA-HV and from 72.5 to 82.2 on \LSATA. The ILP is slightly below the baselines on \textsc{And} (by 0.3 and 0.9 points) and on par on \LSATA~\textsc{Mixed}.

\paragraph{Why local composition falls short} Independent Boolean composition yields zero or multiple valid options in 57.6\% of \LCQA-HV instances and 65.3\% of \LSATA ~instances. Local atomic predictions thus often violate the exactly-one constraint and need an extra resolution rule, whereas the ILP resolves these conflicts jointly under the logical and exactly-one constraints.
% }

\subsection{ Detailed Error Analysis}
\label{app:error-analysis}

We analyze errors at two levels. First, we examine the semantic causes of incorrect atomic judgments, asking why the model assigns high or low confidence to the constructed hypotheses. Second, we examine how the resulting atomic judgments propagate through the logical operators and global inference constraints to produce the final prediction. This separation helps distinguish errors arising from semantic understanding from those arising through logical composition.

\subsubsection{Semantic Causes of Atomic Errors}

Among the inspected \LCQA~errors, we first observe cases in which the model assigns high confidence to incorrect atomics. For example, for the question \textit{What could have a hot handle?}, the model assigns high confidence to \textit{plastic container} and \textit{glass jar}, despite the metal cookware alternatives being more strongly supported by ordinary commonsense. These cases reflect direct errors in atomic scoring.

We also find several cases in which a question admits multiple plausible commonsense interpretations. Open-ended terms such as "might", "could", "where", and "may" encourage the model to consider a broad range of possibilities. In these cases, the model may assign high confidence to atomics outside the gold set, indicating a difference between its interpretation of the question and the interpretation represented by the annotations. Since commonsense judgments can be context-dependent, ambiguity in question interpretation may contribute to some of these errors \citep{palta-etal-2024-plausibly}.

Among the inspected \LSATA~errors, we find several cases in which the model assigns low confidence to labels that are supported by the document. Unlike \LCQA, where multiple commonsense interpretations may be plausible, these errors often involve difficulty identifying all labels that apply to the given paragraph. In particular, the model may recognize the document's general subject while failing to recover a broader or secondary label. This indicates that some atomic errors in \LSATA~result from the model not identifying the full set of applicable labels. 

Across both benchmarks, we also observe cases where the model judges whether an atomic is generally plausible while giving insufficient weight to modifiers or relations expressed in the question. This suggests that some atomic errors arise when contextual conditions are not fully preserved during hypothesis scoring.

Table~\ref{tab:semantic-error-examples} presents representative examples of these semantic error patterns.
\begin{table*}[t]
\centering
\small
\begin{tabular}{p{0.10\linewidth} p{0.27\linewidth} p{0.29\linewidth} p{0.26\linewidth}}
\toprule
\textbf{Benchmark} & \textbf{Input context} & \textbf{Observed atomic behavior} & \textbf{Interpretation} \\
\midrule
LCQA
& What could have a hot handle?
& The model assigns high confidence to \textit{plastic container} and \textit{glass jar}.
& The model incorrectly accepts atomics that are less strongly supported than the metal cookware alternatives. \\
LCQA
& Where do you see tiny bottles of shampoo when away from home?
& The model assigns high confidence to hotels, vacation rentals, cruise shops, and gym showers.
& The open-ended wording may support several possible locations. \\
LCQA
& What is an uncommon side effect of drinking alcohol?
& The model assigns high confidence to \textit{frequent restroom visits}, although the question asks specifically for an uncommon effect.
& The model appears to judge the general plausibility of the effect while giving insufficient weight to the modifier "uncommon". \\
\midrule
LSATA
& A biomedical article on mitochondrial protein import that mentions yeast and mouse.
& The model assigns low confidence to the label \textit{Organisms}.
& The model fails to connect explicit textual evidence to a broader applicable label. \\
LSATA
& An article describing a product launch and providing information about the company.
& The model recognizes the product launch but assigns low confidence to \textit{company description}.
& The model identifies the main event while missing a secondary applicable label. \\
\bottomrule
\end{tabular}
\caption{Representative semantic atomic errors in \LCQA~(LCQA) and \LSATA~(LSATA). The examples include direct atomic scoring errors, broad interpretations of open-ended questions, insufficient attention to modifiers, and failures to identify applicable document labels.}
\label{tab:semantic-error-examples}
\end{table*}

\subsubsection{Logical Propagation of Atomic Errors}

We next examine how incorrect atomic judgments cascade through the logical structure of the options to affect the final prediction. To illustrate this interaction, Table~\ref{tab:logical-propagation} compares instances based on the same question but constructed using different operators, showing how logical composition can tolerate or amplify the underlying scoring errors.

The comparison shows that the same type of atomic error can have different consequences depending on the operator. In the \textsc{or} construction, the false-positive atomics are tolerated because the gold option remains supported by its disjuncts. In the \textsc{and} construction, the additional false-positive atomics support a competing conjunction and lead to an incorrect prediction. In the Mixed construction, the same false positives both invalidate the gold \textsc{neither} option and support a competing \textsc{and} option. 

This behavior is also reflected in the broader error sample. \textsc{and} errors arise when one or both required conjuncts are rejected, whereas \textsc{or} errors occur when the model rejects all atomics that could support the gold option. In contrast, \textsc{neither/nor} errors arise when at least one member of the gold pair is incorrectly accepted. Mixed instances are more complex because one atomic judgment can influence options with different operators.

Because each benchmark instance contains exactly one correct option, global inference enforces the exact-one constraint. However, the model's local evidence scores may make either no option or multiple options logically valid. In these cases, the solver must then adjust some atomic assignments so that exactly one option remains valid. Which assignments change depends on both their confidence scores and their roles across the options. As a result, the final inferred assignments may differ from the model's local preferences even when the logical constraints are correctly enforced.

Overall, compound-level errors usually originate in the model's atomic judgments, while the logical operators and global constraints determine how those errors affect the final prediction.  \textsc{Or} can tolerate some incorrect atomic assignments, whereas \textsc{And} and \textsc{Neither/Nor} may be invalidated by a single error. In \textsc{Mixed} instances, one atomic assignment can influence options governed by different operators, making error propagation more complex.

\begin{table*}[t]
\centering
\small
\begin{tabular}{p{0.08\linewidth} p{0.19\linewidth} p{0.19\linewidth} p{0.23\linewidth} p{0.22\linewidth}}
\toprule
\multicolumn{5}{l}{\textbf{Question:} What could have a hot handle?} \\
\midrule
\textbf{Framing} &
\textbf{Gold option} &
\textbf{Prediction} &
\textbf{Relevant $p^{+}$ scores} &
\textbf{Logical consequence} \\
\midrule
\textsc{Or}
& \textit{metal saucepan} OR \textit{baking tray}
& Gold option
& \textit{metal saucepan}: 0.981; \textit{baking tray}: 0.644; \textit{plastic container}: 0.917; \textit{glass jar}: 0.877
& The gold option remains valid because at least one of its disjuncts is retained. The false-positive atomics do not prevent the correct prediction. \\
\textsc{And}
& \textit{cast iron skillet} AND \textit{baking tray}
& \textit{metal saucepan} AND \textit{plastic container}
& \textit{cast iron skillet}: 0.977; \textit{baking tray}: 0.644; \textit{metal saucepan}: 0.981; \textit{plastic container}: 0.917
& The extra positive atomics support a competing conjunction, causing the gold conjunction to lose under the uniqueness constraint. \\
\textsc{Mixed}
& NEITHER \textit{glass jar} NOR \textit{plastic container}
& \textit{baking tray} AND \textit{plastic container}
& \textit{glass jar}: 0.877; \textit{plastic container}: 0.917; \textit{baking tray}: 0.644
& The false positives invalidate the gold \textsc{Neither/Nor} option, while \textit{plastic container} also supports the competing \textsc{And} option. \\
\bottomrule
\end{tabular}
\caption{Different effects of atomic scoring errors across logical constructions based on the same \LCQA{} question.}
\label{tab:logical-propagation}
\end{table*}

\section{Benchmark Structure and Examples}\label{app:benchmark}

\begin{figure*}[t]
    \centering
    \includegraphics[
        width=\textwidth,
        keepaspectratio
    ]{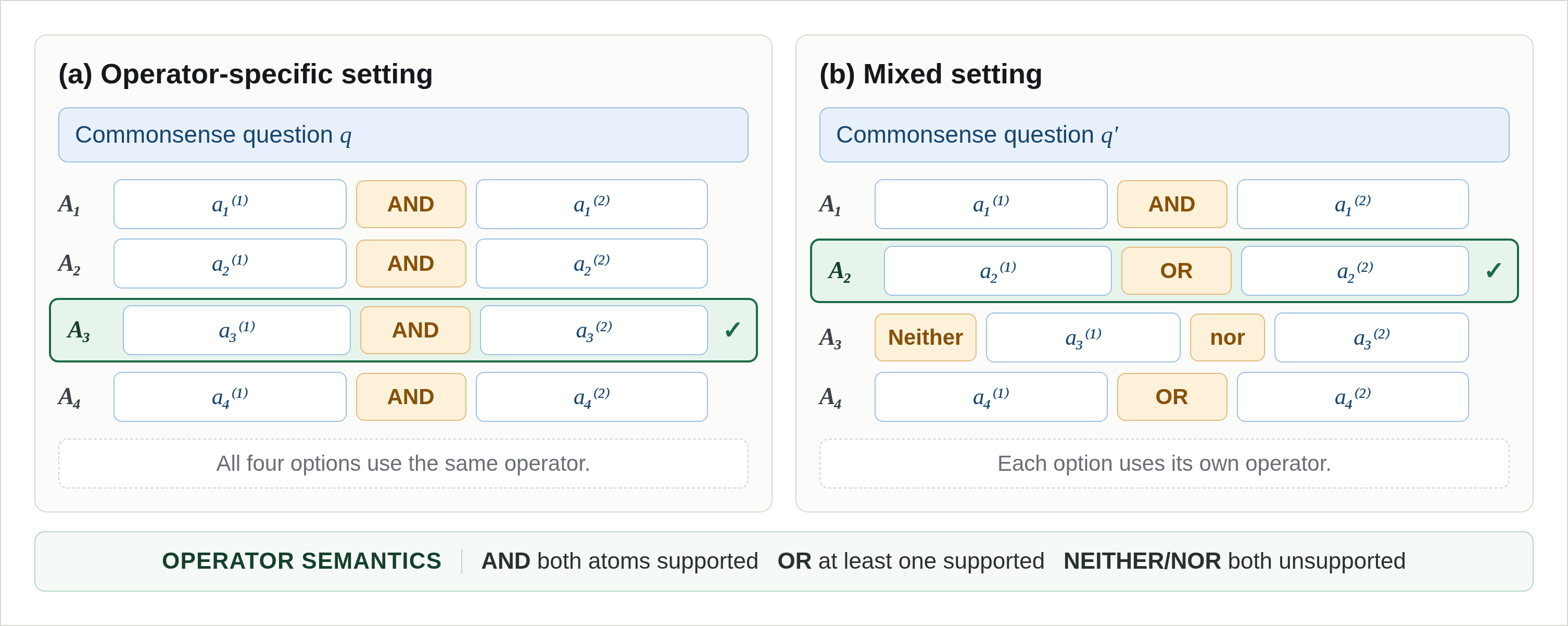}
    \caption{
    Structure of \LCQA~instances. Operator-specific instances use the same operator across all four options, whereas \textsc{Mixed} instances may contain different operators.
    }
    \label{fig:lcsqa_structure}
\end{figure*}

\begin{figure*}[t]
    \centering
    \includegraphics[
        width=\textwidth,
        height=0.25\textheight,
        keepaspectratio
    ]{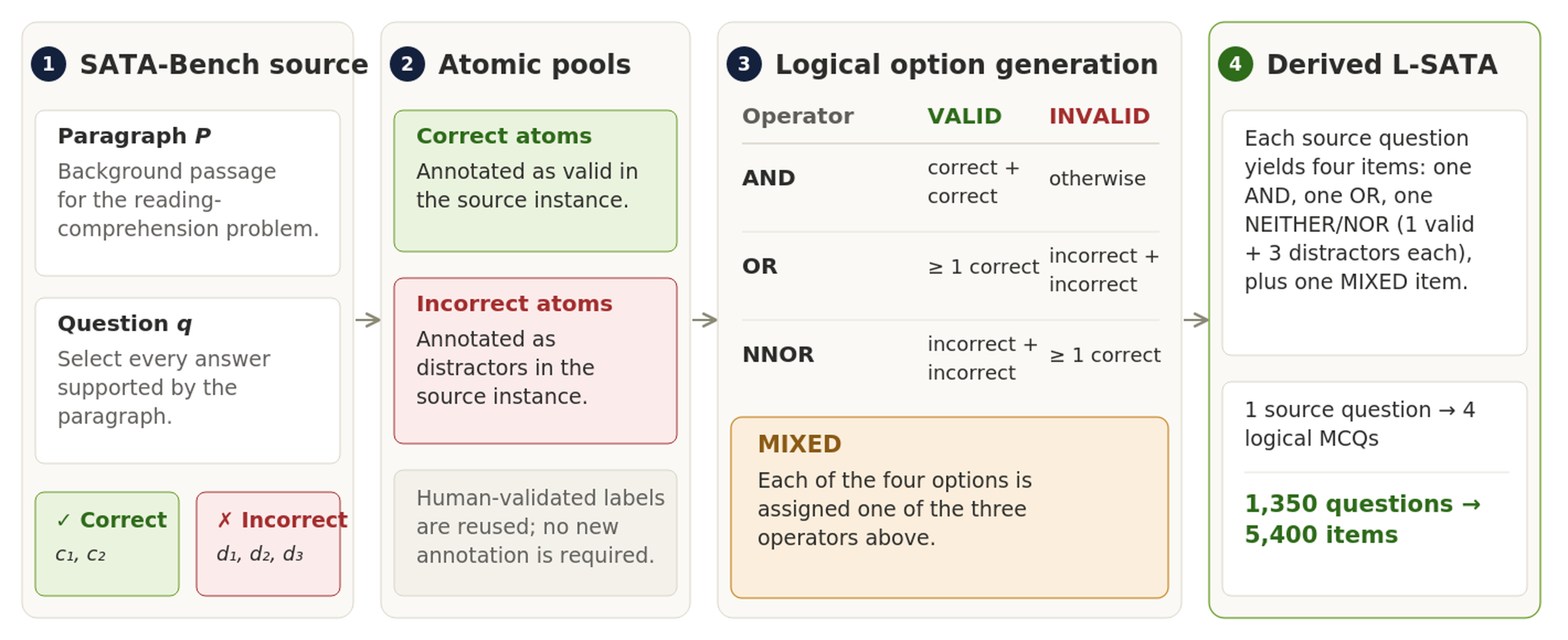}
    \caption{
    Construction of \LSATA~from SATA-Bench. Each source instance provides a
    paragraph, a reading-comprehension question, and independently annotated
    correct and incorrect atomic answers. Pairs of atomic answers are
    combined according to the semantics of \textsc{And}, \textsc{Or}, and
    \textsc{Neither/Nor}. Each source question produces one item for each
    operator-specific setting and one \textsc{Mixed} item.
    }
    \label{fig:lsata_construction}
\end{figure*}

Figures~\ref{fig:lcsqa_structure} and~\ref{fig:lsata_construction} present the structure and construction of the benchmarks. %The values in parentheses are ground-truth binary labels for the atomic answers.

%\section{Representative Benchmark Examples}
%\label{app:benchmark-examples}

Figures~\ref{fig:lcqa-examples} and~\ref{fig:lsata-examples} present representative instances from the four logical settings. The values in parentheses are ground-truth binary labels for the atomic answers.

\begin{figure*}[p]
\centering

\begin{tabularx}{\textwidth}{@{}Y@{\hspace{3mm}}Y@{}}

\begin{benchmarkcard}{QID 133: \textsc{And}}
\footnotesize

\textbf{Question.}
Helium, magnesium, and sulfur are likely to be found where in a school?

\medskip

\benchmarkoption{A}{And}{art studio (0)}{physical education (0)}
\goldbenchmarkoption{B}{And}{science lab (1)}{materials workshop (1)}
\benchmarkoption{C}{And}{materials workshop (1)}{art studio (0)}
\benchmarkoption{D}{And}{science lab (1)}{physical education (0)}

\medskip
\textbf{Gold composition:}
\[
1 \land 1 = 1
\]
\end{benchmarkcard}

&

\begin{benchmarkcard}{QID 604: \textsc{Or}}
\footnotesize

\textbf{Question.}
What do horses do to get energy?

\medskip

\goldbenchmarkoption{A}{Or}{graze on pasture (1)}{rest in stables (0)}
\benchmarkoption{B}{Or}{play in fields (0)}{nibble on flowers (0)}
\benchmarkoption{C}{Or}{play in fields (0)}{rest in stables (0)}
\benchmarkoption{D}{Or}{nibble on flowers (0)}{rest in stables (0)}

\medskip
\textbf{Gold composition:}
\[
1 \lor 0 = 1
\]
\end{benchmarkcard}

\\[3mm]

\begin{benchmarkcard}{QID 1223: \textsc{Neither/Nor}}
\footnotesize

\textbf{Question.}
Where is confetti thrown from the rooftops?

\medskip

\benchmarkoptionNN{A}{festive parades (1)}{isolated parks (0)}
\benchmarkoptionNN{B}{public events (1)}{business functions (0)}
\benchmarkoptionNN{C}{public events (1)}{isolated parks (0)}
\goldbenchmarkoptionNN{D}{indoor gatherings (0)}{isolated parks (0)}

\medskip
\textbf{Gold composition:}
\[
\neg 0 \land \neg 0 = 1
\]
\end{benchmarkcard}

&

\begin{benchmarkcard}{QID 1666: \textsc{Mixed}}
\footnotesize

\textbf{Question.}
Sarah took poison by accident. She found it in the cabinet and thought
that it was what?

\medskip

\benchmarkoption{A}{And}{vitamin energy tonic (0)}{sports hydration drink (0)}
\benchmarkoption{B}{And}{herbal cold remedy (1)}{sports hydration drink (0)}
\goldbenchmarkoption{C}{Or}{herbal cold remedy (1)}{sports hydration drink (0)}
\benchmarkoption{D}{And}{child's flavored syrup (1)}{sports hydration drink (0)}

\medskip
\textbf{Gold composition:}
\[
1 \lor 0 = 1
\]

The same $(1/0)$ pattern is invalid under \textsc{And}.
\end{benchmarkcard}

\end{tabularx}

\caption{
Representative \LCQA{} instances covering the \textsc{And},
\textsc{Or}, \textsc{Neither/Nor}, and \textsc{Mixed} settings.
Values in parentheses denote ground-truth binary labels for the atomic
answers. The highlighted option is the unique gold option.
}
\label{fig:lcqa-examples}
\end{figure*}

\begin{figure*}[p]
\centering

\begin{tabularx}{\textwidth}{@{}Y@{\hspace{3mm}}Y@{}}

\begin{benchmarkcard}{QID 718: \textsc{And}}
\scriptsize

\textbf{Source domain.}
Biomedical MeSH classification

\medskip

\textbf{Article.}
Role of LRRK2 in the regulation of dopamine receptor trafficking.

\medskip

\textbf{Question.}
Given the above article, which MeSH root categories can be assigned to it?

\medskip

\benchmarkoption{A}{And}{Geographicals (0)}{Disciplines and Occupations (0)}
\benchmarkoption{B}{And}{Anthropology, Education, Sociology, and Social Phenomena (0)}{Technology, Industry, and Agriculture (0)}
\goldbenchmarkoption{C}{And}{Chemicals and Drugs (1)}{Diseases (1)}
\benchmarkoption{D}{And}{Geographicals (0)}{Named Groups (0)}

\medskip
\textbf{Gold composition:}
\[
1 \land 1 = 1
\]
\end{benchmarkcard}

&

\begin{benchmarkcard}{QID 158: \textsc{Or}}
\scriptsize

\textbf{Source domain.}
Science reading comprehension

\medskip

\textbf{Passage excerpt.}
Magnets exert forces on certain materials. Unlike ordinary contact
forces, magnetic forces can act without the magnet touching the object.

\medskip

\textbf{Question.}
How are magnetic forces different from other forces?

\medskip

\benchmarkoption{A}{Or}{Other forces cause a change called motion (0)}{Magnets produce force by touching (0)}
\benchmarkoption{B}{Or}{All forces are essentially the same because they use touching (0)}{Other forces cause a change called motion (0)}
\benchmarkoption{C}{Or}{All forces are essentially the same because they use touching (0)}{Magnets produce force by touching (0)}
\goldbenchmarkoption{D}{Or}{Magnetic force can produce motion without actual contact with certain materials (1)}{Other forces cause a change called motion (0)}

\medskip
\textbf{Gold composition:}
\[
1 \lor 0 = 1
\]
\end{benchmarkcard}

\\[3mm]

\begin{benchmarkcard}{QID 96: \textsc{Neither/Nor}}
\scriptsize

\textbf{Source domain.}
Literary reading comprehension

\medskip

\textbf{Passage excerpt.}
``Femme'' describes a dangerously attractive woman who uses her sexual
attractiveness to lure men to their downfall. Madame White Snake is
depicted as bewitchingly beautiful.

\medskip

\textbf{Question.}
What does ``femme'' refer to?

\medskip

\benchmarkoptionNN{A}{a bewitchingly beautiful woman (1)}{all women (0)}
\goldbenchmarkoptionNN{B}{all women (0)}{a true lover (0)}
\benchmarkoptionNN{C}{a sexually seductive woman (1)}{a true lover (0)}
\benchmarkoptionNN{D}{a sexually seductive woman (1)}{a bewitchingly beautiful woman (1)}

\medskip
\textbf{Gold composition:}
\[
\neg 0 \land \neg 0 = 1
\]
\end{benchmarkcard}

&

\begin{benchmarkcard}{QID 1095: \textsc{Mixed}}
\scriptsize

\textbf{Source domain.}
EU legal-document classification

\medskip

\textbf{Document.}
Council Decision of 7 March 2011 appointing members and alternates of the
Advisory Committee for the Coordination of Social Security Systems for
Ireland, France, and the Netherlands.

\medskip

\textbf{Question.}
What concepts does the above document include?

\medskip

\benchmarkoptionNN{A}{advisory committee (EU) (1)}{Maghreb (0)}
\goldbenchmarkoption{B}{Or}{advisory committee (EU) (1)}{social security (1)}
\benchmarkoptionNN{C}{advisory committee (EU) (1)}{social security (1)}
\benchmarkoption{D}{Or}{special polymer (0)}{uncultivated land (0)}

\medskip
\textbf{Gold composition:}
\[
1 \lor 1 = 1
\]

Option C contains the same atomics but is invalid under
\textsc{Neither/Nor}:
\[
\neg 1 \land \neg 1 = 0.
\]
\end{benchmarkcard}

\end{tabularx}

\caption{
Representative \LSATA{} instances from four source domains, covering the
\textsc{And}, \textsc{Or}, \textsc{Neither/Nor}, and \textsc{Mixed}
settings. Values in parentheses denote ground-truth binary labels for the
atomic answers. The highlighted option is the unique gold option. Passage
and document excerpts are shortened for readability.
}
\label{fig:lsata-examples}
\end{figure*}

\subsection{Structural Shortcuts}
\label{app:structural-shortcuts}
% \new{
Because exactly one option is valid by construction, some instances could be solved from the option structure alone. To quantify this, we remove all semantic information and exhaustively enumerate every true/false assignment to the atomic answers of an instance. We call an instance \emph{structurally forced} if, under the exactly-one constraint, every feasible assignment selects the same option.

Only 4.30\% of \LCQA-HV and 5.15\% of \LSATA~instances are structurally forced. These are concentrated in \textsc{Or} (42/43 and 98/103, respectively), because a true atom shared across \textsc{Or} options activates several options at once, so the exactly-one constraint rules out most assignments. None of the \textsc{And} or \textsc{Neither/Nor} instances are forced, and only 1 and 5 \textsc{Mixed} instances across both datasets respectively are forced. Thus, structure alone does not determine the answer for more than 94\% of instances, so correct atomic judgments remain necessary. Atomic accuracy is 83.0\% on \LCQA-HV and 82.4\% on \LSATA.
% }
\section{Prompt Templates}
\label{app:prompts}

We present prompt templates used in our experiments. Each template retains the task description, instance-specific inputs, principal instructions, and required output format.

\subsection{Hypothesis Construction}
\label{app:hypothesis-prompts}
The constraint specifies the requirements expressed by the question, against which each atomic answer is evaluated.
For each atomic answer, we construct a fixed positive hypothesis $h_C^{+}(a)$, stating that the atomic answer satisfies the constraint, and a negative hypothesis $h_C^{-}(a)$, stating that it does not.

\subsection{Paired Multiple-Choice Confidence}
\label{app:paired-mc-prompt}

The paired multiple-choice prompt presents the positive and negative hypotheses as options A and B within a single prompt. The model selects the interpretation that is correct with respect to the context and constraint.

\paragraph{\LCQA}

\begin{quote}
\small\ttfamily
You evaluate which of two competing hypotheses about an atomic answer is correct with respect to a required commonsense constraint.

\textbf{Input}

Question: \{\{question\}\}

Constraint: \{\{constraint\}\}

Atomic statement: \{\{atomic\_statement\}\}

Option A: \{\{H\_plus\}\}\\
Option B: \{\{H\_minus\}\}

\textbf{Instructions}

\begin{itemize}
\setlength\itemsep{0pt}
\item Judge whether the atomic answer fulfills that requirement.
\item Judge the atomic answer independently of the other answer options.
\item Select exactly one option.
\end{itemize}

\textbf{Output}

Return only the single letter \texttt{A} or \texttt{B}.

Answer:
\end{quote}

\paragraph{\LSATA}

\begin{quote}
\small\ttfamily
You evaluate which of two competing hypotheses about an atomic answer is correct with respect to a required reading-comprehension constraint.

\textbf{Input}

Passage: \{\{paragraph\}\}

Question: \{\{question\}\}

Constraint: \{\{constraint\}\}

Atomic statement: \{\{atomic\_statement\}\}

Option A: \{\{H\_plus\}\}\\
Option B: \{\{H\_minus\}\}

\textbf{Instructions}

\begin{itemize}
\setlength\itemsep{0pt}
\item Determine whether the atomic answer fulfills that requirement as described in the passage.
\item Judge the atomic answer independently of the other answer options.
\item Select exactly one option.
\end{itemize}

\textbf{Output}

Return only the single letter \texttt{A} or \texttt{B}.

Answer:
\end{quote}

\subsection{Independent True--False Confidence}
\label{app:independent-prompt}

Independent true--false confidence evaluates the positive and negative hypotheses in separate model calls.

\paragraph{\LCQA}

\begin{quote}
\small\ttfamily
You determine whether a hypothesis about an atomic answer is true with respect to a required commonsense constraint.

\textbf{Input}

Question: \{\{question\}\}

Constraint: \{\{constraint\}\}

Atomic statement: \{\{atomic\_statement\}\}

Hypothesis: \{\{H\_plus\}\} or \{\{H\_minus\}\}

\textbf{Instructions}

\begin{itemize}
\setlength\itemsep{0pt}
\item Determine whether the supplied hypothesis is true.
\item Judge the atomic answer against the constraint on its own terms.
\end{itemize}

\textbf{Output}

Return only \texttt{True} or \texttt{False}.

Answer:
\end{quote}

\paragraph{\LSATA}

\begin{quote}
\small\ttfamily
You determine whether a hypothesis about an atomic answer is true with respect to a reading-comprehension constraint.

\textbf{Input}

Passage: \{\{paragraph\}\}

Question: \{\{question\}\}

Constraint: \{\{constraint\}\}

Atomic statement: \{\{atomic\_statement\}\}

Hypothesis: \{\{H\_plus\}\} or \{\{H\_minus\}\}

\textbf{Instructions}

\begin{itemize}
\setlength\itemsep{0pt}
\item Determine whether the supplied hypothesis is true according to the passage.
\item Judge the atomic answer against the passage and constraint on its own terms.
\end{itemize}

\textbf{Output}

Return only \texttt{True} or \texttt{False}.

Answer:
\end{quote}

Separate calls are made for $h_C^{+}(a)$ and $h_C^{-}(a)$.

\subsection{Generation Sampling}
\label{app:generation-sampling-prompt}

Generation sampling uses the same benchmark-specific templates as paired multiple-choice confidence.

\begin{quote}
\small\ttfamily
You evaluate which of two competing hypotheses about an atomic answer is correct with respect to the required question constraints.

\textbf{Input}

[Passage: \{\{paragraph\}\}]

Question: \{\{question\}\}

Constraint: \{\{constraint\}\}

Atomic statement: \{\{atomic\_statement\}\}

Option A: \{\{H\_plus\}\}\\
Option B: \{\{H\_minus\}\}

\textbf{Instructions}

\begin{itemize}
\setlength\itemsep{0pt}
\item Judge whether the atomic answer fulfills the specific requirement expressed by the constraint.
\item Judge the atomic answer independently of the other answer options.
\item Select exactly one option.
\end{itemize}

\textbf{Output}

Return only the single letter \texttt{A} or \texttt{B}.

Answer:
\end{quote}

The passage field is included for \LSATA~and omitted for \LCQA. Rather than extracting answer-token probabilities, we sample five responses and compute the positive and negative evidence scores from the empirical frequencies of \texttt{A} and \texttt{B}.

\subsection{Verbalized Confidence}
\label{app:verbalized-prompt}

Verbalized confidence evaluates the positive and negative hypotheses in separate calls and asks the model to report a numerical confidence value.

\paragraph{\LCQA}

\begin{quote}
\small\ttfamily
You report your confidence that a hypothesis about an atomic answer is true with respect to the required commonsense question constraints.

\textbf{Input}

Question: \{\{question\}\}

Constraint: \{\{constraint\}\}

Atomic statement: \{\{atomic\_statement\}\}

Hypothesis: \{\{H\_plus\}\} or \{\{H\_minus\}\}

\textbf{Instructions}

\begin{itemize}
\setlength\itemsep{0pt}
\item Evaluate whether the supplied hypothesis is true.
\item Judge the atomic answer against the constraint on its own terms.
\end{itemize}

\textbf{Output}

Return one integer from $0$ to $10$, where $0$ indicates complete confidence that the hypothesis is false and $10$ indicates complete confidence that it is true.

Confidence:
\end{quote}

\paragraph{\LSATA}

\begin{quote}
\small\ttfamily
You report your confidence that a hypothesis about an atomic answer is true with respect to a reading-comprehension question constraint.

\textbf{Input}

Passage: \{\{paragraph\}\}

Question: \{\{question\}\}

Constraint: \{\{constraint\}\}

Atomic statement: \{\{atomic\_statement\}\}

Hypothesis: \{\{H\_plus\}\} or \{\{H\_minus\}\}

\textbf{Instructions}

\begin{itemize}
\setlength\itemsep{0pt}
\item Evaluate whether the supplied hypothesis is true according to the passage.
\item Judge the atomic answer against the passage and constraint on its own terms.
\end{itemize}

\textbf{Output}

Return one integer from $0$ to $10$, where $0$ indicates complete confidence that the hypothesis is false and $10$ indicates complete confidence that it is true.

Confidence:
\end{quote}

Separate calls are made for $h_C^{+}(a)$ and $h_C^{-}(a)$.

\subsection{Representative Demonstrations}
\label{app:prompt-demonstrations}

The full prompts contain fixed demonstrations. We show one representative example from each benchmark.

\paragraph{\LCQA}

\begin{quote}
\small\ttfamily
Question: Where can you see a mountain in your own home?

Constraint: Someone can see a mountain from inside their home

Atomic statement: a window facing the mountains

Option A: The atomic statement satisfies the constraint.

Option B: The atomic statement fails the constraint.

Answer: A
\end{quote}

\paragraph{\LSATA}

\begin{quote}
\small\ttfamily
Passage: The Legal Aid Society was barred from returning
to its headquarters near the World Trade Center site
because of environmental concerns.

Question: Why could the Legal Aid Society not return
to its original headquarters?

Constraint: The reason Legal Aid Society could not return
to its original headquarters

Atomic statement: contamination risk from the nearby
World Trade Center site

Option A: The atomic statement satisfies the constraint.

Option B: The atomic statement fails the constraint.

Answer: A
\end{quote}

\subsection{Direct Prompting}
\label{app:direct-prompts}
% \new{
Direct prompting presents the four options as a multiple-choice question, and the prediction is read out from the first-token log-probabilities of the letters A--D. The few-shot variants prepend $n$ fixed demonstrations from the instance's operator setting. Zero-shot chain-of-thought asks the model to reason step by step and end with ``FINAL ANSWER: X''. We parse the answer letter from the generated response and issue one deterministic repair request if parsing fails. Responses that remain unparseable are counted as incorrect.
% }
\paragraph{\LCQA}

\begin{quote}
\small\ttfamily
Answer the following commonsense question by selecting the correct option.

\textbf{Input}

[Demonstrations (for 1-, 2- and 3-shot only)]

Question: \{\{question\}\}

A. \{\{A\}\}

B. \{\{B\}\}

C. \{\{C\}\}

D. \{\{D\}\}

\textbf{Instructions}

\begin{itemize}
\setlength\itemsep{0pt}
\item Select the correct option.
\item Respond with only the single capital letter A, B, C, or D.
\item Do not include explanation or punctuation.
\end{itemize}

\textbf{Output}

Return only the single letter A, B, C, or D.

Answer:
\end{quote}

\paragraph{\LSATA}

\begin{quote}
\small\ttfamily
Answer the following reading-comprehension question by selecting the correct option.

\textbf{Input}

[Demonstrations (for 1-, 2- and 3-shot only)]

Passage: \{\{paragraph\}\}

Question: \{\{question\}\}

(A) \{\{A\}\}

(B) \{\{B\}\}

(C) \{\{C\}\}

(D) \{\{D\}\}

\textbf{Instructions}

\begin{itemize}
\setlength\itemsep{0pt}
\item Select the correct option.
\item Respond with only the single capital letter A, B, C, or D.
\item Do not include explanation or punctuation.
\end{itemize}

\textbf{Output}

Return only the single letter A, B, C, or D.

Answer:
\end{quote}

\end{document}